\documentclass[a4paper]{article}
\pdfoutput=1

\usepackage{amsmath}

\usepackage{amssymb}

\usepackage[figuresright]{rotating}
\usepackage{paralist}
\usepackage{graphicx}
\usepackage{pgfplots}
\usepackage{pgfplotstable}
\usepackage{tikz}
\usepackage{scalefnt}
\usepackage[hidelinks,pdftex]{hyperref}
\usepackage{xspace}
\usepackage{tabularx}
\usepackage{caption}
\usepackage{subcaption}
\usepackage{booktabs}
\usepackage{breqn}
\usepackage{epstopdf}
\usepackage{siunitx}
\usepackage{url}
\usepackage{authblk}

\DeclareMathOperator*{\argmax}{arg\,max}
\DeclareMathOperator*{\fscore}{\textit{F}\text{-score}}
\DeclareMathOperator*{\precis}{\text{Precision}}
\DeclareMathOperator*{\recall}{\text{Recall}}

\newcommand{\wisp}{W\textsuperscript{2}ISP}
\newcommand{\fsc}{\textit{F}-score\xspace}
\newcommand{\eg}{e.g.\xspace}
\newcommand{\ie}{i.e.\xspace}

		\title{Learning from Imbalanced Multiclass Sequential Data Streams Using Dynamically Weighted Conditional Random Fields}
		
		\author[1]{Roberto L. {Shinmoto Torres}\thanks{Corresponding author}}
		\author[1]{Damith C. Ranasinghe} 
		\author[2]{Qinfeng Shi}
		\author[2]{Anton van den Hengel}
		\affil[1]{Auto-ID Lab, The School of Computer Science,The University of Adelaide, Adelaide, SA 5005, Australia}
	    \affil[2]{The Australian Centre for Visual Technologies ({ACVT}),  The School of Computer Science, The University of Adelaide, Adelaide, SA 5005, Australia}

\begin{document}

%
%

\maketitle
		
		\begin{abstract}
		The present study introduces a method for improving the classification performance of imbalanced multiclass data streams from wireless body worn sensors. 
		Data imbalance is an inherent problem in activity recognition caused by the irregular time distribution of activities, which are sequential and dependent on previous movements. 
		We use conditional random fields (CRF), a graphical model for structured classification, to take advantage of dependencies between activities in a sequence. 
		However, CRFs do not consider the negative effects of class imbalance during training. 
		We propose a class-wise dynamically weighted CRF (dWCRF) where weights are automatically determined during training by maximizing the expected overall F-score. 
		Our results based on three case studies from a healthcare application using a batteryless body worn sensor, demonstrate that our method, in general, improves overall and minority class F-score when compared to other CRF based classifiers and achieves similar or better overall and class-wise performance when compared to SVM based classifiers under conditions of limited training data. 
		We also confirm the performance of our approach using an additional battery powered body worn sensor dataset, achieving similar results in cases of high class imbalance. 
		\end{abstract}



\section{Introduction}

\label{sec:intro}

Developments in emerging wireless sensor technologies is enabling a multitude of applications, particularly in healthcare practice for applications such as location tracking, medical monitoring of patients and recognition of performed activities. 
The recognition of activities in older people is of particular interest as a means of preventing injuries from events such as falls by providing an early intervention, 
or identification of function decline, as in those with Alzheimer's or Parkinson's disease, to enact preventive interventions.

One of the main challenges in human activity recognition is the fact that sensor data is usually imbalanced as data from all possible activities are not necessarily equally distributed. 
This is because people naturally perform some activities that are of longer duration than others. For instance, in the context of patient monitoring in hospitals or nursing homes, resting activities such as lying on bed (\ie sleeping) or sitting on a chair or on the bed are of longer duration than ambulating activities, where destinations, for example  a rest room, are very close. 
Furthermore, data is more easily collected for certain activities than others; for instance, data from activities performed closer to the sensing infrastructure (\eg\ motion sensors on ceilings) can be more easily collected than those activities performed farther from the sensors.  

This paper presents a novel method for learning from imbalanced data using conditional random fields. 
We propose a class-wise cost parameter based classifier, that is able to consider the dependencies between activities. 
The classifier considers the influence of individual classes for learning from sequential imbalanced multiclass datasets. 
The cost parameters (weights) are not fixed as they are dynamically adjusted during the training process; while the classifier seeks to optimize the model's expected overall \fsc to minimize both false positives (false alarms) and false negatives (missed classifications). 
The performance of our approach is evaluated in three case studies in the context of recognizing activities of older people instrumented with a single body worn batteryless sensor. 
The results are validated with an external dataset where levels of imbalance were incremented.

\subsection{Scope and Background}

Imbalanced data can negatively affect the training of machine learning algorithms as the classifier can be biased to prefer the majority class \cite{He_5128907,Japkowicz2002}. 
This can be a serious problem as minority classes are of great importance in applications such as human activity monitoring.  For example, older people previously assessed as being at risk of falling performing a short duration ambulation---as opposed to large amounts of time spent in resting postures such as lying in bed---are potentially at a risk of falling and injury \cite{Rapp2012}. 
Hence it is important to increase the overall classification performance and, in particular, the classifier performance in identifying minority classes such as ambulation in our example.

Another challenge arising from the nature of human activity recognition problems is the difficulty of collecting and labelling large datasets from activities of daily living (ADL). This is indeed the case for applications such as monitoring patients in acute hospitals where collecting data to learn activities of hospitalized older people is very difficult due to physical limitations resulting from their older age and associated ailments\cite{Chernbumroong20131662,Ransinghe:Advances2013}. 
Therefore, a classifier which is highly accurate at predicting all activity classes in datasets where availability of training data is scarce is highly desirable. 

Given that human activities are sequential and a person can perform the same activity in different manners; we base our classifier in conditional random fields (CRFs)~\cite{Lafferty2001}, a graphical model for structured classification that captures dependency relationships between performed activities as described by sensor observations. 
We evaluate our proposed classifier with three case studies from healthy and hospitalized older people using a battery-less body worn sensor where the data streams from the wearable sensors are irregular and sparse, noisy and class imbalanced. 
We confirm our findings using a publicly available dataset for human activity recognition using battery powered body worn sensors; in this scenario we modified the imbalance to levels similar or higher than those of our case studies.

The study presented in this article is part of our ongoing research aimed at recognizing activities 
by older people 
in hospital and nursing home settings for falls prevention, as described in \cite{Visvanathan_6347326,Ranasinghe2014,Shinmoto:6548154}. This paper presents a method for learning from imbalanced sensor data streams, with limited training data, using conditional random fields.

\subsection{Related works}
\label{sec:related}

This section reviews previous methods developed for improving the classification of imbalanced data, 
such as data re-sampling \cite{Batista:2004,Tahir20123738,Das_2014,Xue-6906278}, adjusting decision thresholds \cite{Maloof2003learning} or the inclusion of cost parameters or weights into the classification algorithm \cite{Domingos:1999:Metacost,Liu:4053137,Zhou:COIN358,Pang201387,Gama2003417,Jiang2014211,Huang_1527706,Lannoy_6036156}. 
The approach presented in this article is based on the latter.  

The main issue with re-sampling techniques \cite{Batista:2004,Tahir20123738,Das_2014} is that the removal or introduction of data can modify the sequence structure and its meaning. 
This is an issue in some real world applications that require maintaining the original data structure. 
For example, modifying parts of a sentence for text classification can effectively change the meaning of the message. 
Similarly, in human activity recognition the time sequence is important to determine the flow of movement or activities and introducing data can change the sequence of activities and the way it is analyzed, \eg\ affecting transition probabilities in Markov chains.

Decision threshold methods such as that of \cite{Maloof2003learning} achieved similar results to re-sampling techniques and used receiver operating characteristic (ROC) curves to decide which decision threshold produces the best performance. 
However, ROC curves 
depend on measuring specificity which does not reflect the errors in imbalanced data; this is due to specificity of the minority class being conditioned to its true negative measurement which includes the true positives of the majority class and thus leading to over optimistic results. 

In the case of cost parameter methods, these require the inclusion of fixed class-wise costs into the objective function of the classifier during training to reinforce the learning of the under-represented classes. 
Generally, cost sensitive learning approaches have been reported to perform better than re-sampling techniques in some applications \cite{He_5128907}.  
Some cost parameters have the form of a cost matrix that weighs each possible misclassification case, giving higher costs to misclassifications of a minority class observation in comparison to majority classes \cite{Domingos:1999:Metacost,Liu:4053137,Zhou:COIN358}. 
Costs have also been used to rescale the data. 
This is done by re-weighting or re-sampling the training samples; or moving decision thresholds according to their costs.  
These costs are usually user provided, \eg\ from a cost matrix; these methods are reported to work well in binary data and only in some multiclass cases \cite{Zhou:COIN358}. 
Moreover, these costs are fixed during training, but in real world applications ---as is our case--- costs can change for various reasons\cite{Zhou:COIN358}.

The method of Huang et al.~\cite{Huang_1527706} introduced a fixed set of weights for each class for a binary SVM algorithm. 
The binary weights ratio was inversely proportional to their respective class population ratio in the training data. 
This achieved a marginal improvement for the minority class accuracy at the cost of possible overall accuracy reduction. 
In Jiang et al.~\cite{Jiang2014211}, weights calculated from the misclassification cost of each class were introduced into Bayesian network classifiers.

The study of Gimpel et al.~\cite{Gimpel2010}, focused on improving the classification performance by modifying costs according to specific performance tasks (task-wise) such as improving recall, precision or both (as in \fsc) \cite{Gimpel2010} as opposed to classification error minimization as in previous studies. 
The method itself did not consider class specific parameters and parameter calculations required an extensive validation process as parameters were not learned in training. 
The introduction of weights in CRF (WCRF) is not new; however, previous approaches only considered using a fixed set of weights during training for optimization 
\cite{Lannoy_6036156}; however, finding an optimal set of weights \cite{Gimpel2010,Jiang2014211} require an extensive validation process. 

These previously mentioned methods \cite{Jiang2014211,Lannoy_6036156,Gimpel2010} require empirical calculation of parameters. 
This process can be cumbersome and computationally expensive. For example, in an extensive grid search for suitable weights, the number of validation operations is of the form $V^{\mathcal{M}}$, where $V$ is the cardinality of the parameters' value range and $\mathcal{M}$ the number of parameters. 
In addition, objective function optimization based on classification error (1-accuracy) minimization is not suitable for imbalanced data. 
This is because the resulting measure, accuracy, is largely favoured by the dominant class and does not provide performance information regarding the predicted minority class~\cite{Chawla_Data_mining2005}. 

Other studies, such as that of Gama et al.~\cite{Gama2003417} dynamically changed the cost of the naive Bayes classifier by verifying and introducing classification errors on each iteration; this method optimized the squared loss function as opposed to the 0--1 loss function. 
However, this study was only compared to the performance of a naive Bayes approach. 
In Pang et al.~\cite{Pang201387}, dynamically learned parameters were introduced into a binary linear proximal SVM (LPSVM) objective function where weights were proportional to the ratio of the other class population. 
The study of Pletscher et al.~\cite{Pletscher_2010} presented a method for generalizing structured classifiers such as CRF and Structured SVM (SSVM) and introduced a cost parameter given by the Hamming distance between predicted output and ground truth which was dynamically calculated during training. 
Although the methods in \cite{Pang201387,Pletscher_2010} presented dynamically calculated cost parameters, the model optimization was based on classification error minimization. 
Other studies offered alternative methods \cite{Soda20111801,Beyan20151653}. 
The method of Soda \cite{Soda20111801} decided between an unbalanced or a balanced classifier for every observation and measured its performance based on accuracy; while Beyan et al. \cite{Beyan20151653} proposed a hierarchical method based on clustering and outlier detection that, in general, was not significantly better than other methods. Moreover, both studies did not consider multi-class problems.

The study of Dimitroff et al.~\cite{Dimitroff2014} considered a learning algorithm for binary classification using a maximum likelihood model (weighted maximum entropy) to optimize the expected \fsc during training where weights were calculated autonomously. 
Our study extends weighted maximum entropy \cite{Dimitroff2014}, originally proposed for binary classification problems, to structured prediction using CRF. 
We use CRF in order to model the dependency between consecutive activities described by sensor observations in the training sequences and applied in previous research  \cite{Shinmoto:6548154,ShinmotoTorres2013}.

In summary, most real-life data is imbalanced and methods that modify the data structure can also change the meaning of the events being studied, specially in the case of sequential data. 
Similarly, the use of optimization metrics such as accuracy are affected by the majority class performance. 
In the case of cost sensitive learning methods, most studies used a fixed set of weights where obtaining an optimal set requires a large amount of validation processes. 
In other dynamically calculated weights methods, the models were optimized based on accuracy. 
Our proposed method is an alternative that considers dynamically calculated weights that optimize the overall F-score to reduce false positives and false negatives for multi-class structured prediction using CRF.

\subsection{Paper Contributions}

Our approach extends existing knowledge by solving current limitations: degradation of sequential data by re-sampling; learning models controlled by measures (accuracy and specificity) biased towards the dominant class; and inclusion of class specific parameters that are fixed or require extensive validation or optimization. 
We make the following contributions: 

\begin{enumerate} 
    \item We present a novel cost sensitive learning method for imbalanced multiclass data classification in real time, based on weighted conditional random fields (WCRF) where the optimization process is based on maximizing the expected overall \fsc. In this method, the cost parameters are dynamically computed during training. To our knowledge, this is the first attempt for multiclass classifier optimization based on \fsc metric to learn from imbalanced data where the cost parameters are also learned, in particular, for graphical models such as CRF. 
    \item We apply our method to two scenarios; the first considers three case studies of sequential data streams from healthy and hospitalized older people using a batteryless body worn sensor over their clothing in clinical and hospital settings; and we generalize our results with a second scenario that considers a battery powered body worn sensor dataset with increased imbalance. 
    \item We achieve better overall performance 
    in comparison to other CRF based classifiers and similar (p-value $> 0.44$) 
    or higher performance than other support vector machine (SVM) based methods in the context of scarce training data, which is often the case with practical applications using supervised methods to obtain highly accurate predictive models. 
\end{enumerate} 

\section[Proposed Dynamically Weighted Learning Method]{\texorpdfstring{Proposed Dynamically Weighted Learning\\ Method}{Proposed Dynamically Weighted Learning Method}}
\label{sec:methods}

\subsection{Background}
\label{ssec:CRF}
In this section we briefly revisit CRFs, a probabilistic graphical model for structured classification \cite{Lafferty2001, Sutton2006}, as background to defining our method in Section~\ref{ssec:WCRF}. 

Let us assume a sequence of input observations $\{x_t\}_{t=1}^{T}$ and their corresponding labels $\{y_t\}_{t=1}^{T}$, where $y_t \in \{1\cdots K\}$ and $K$ is the number of classes to infer. 
The advantage of CRF is that it constructs pairwise relationships between adjacent hidden variables and their corresponding observations, a property from the first order Markov assumption. 
The probability distribution in CRF is given by the conditional probability defined as
\begin{align}
p(y|x,\lambda)&=\frac{1}{Z}\exp \left( \sum_{}^{T}\phi_t \left(y_{t-1},y_t,x;\lambda\right)\right), \\
Z(x) &= \sum_{y_{1\cdots T}} \exp \left( \sum_{}^{T} \phi(y_{t-1},y_t,x;\lambda) \right).
\end{align}
Here the potential function $\exp(\phi(y_{t-1},y_t,x;\lambda))$ follows the logistic model function 
\begin{equation}
\phi(y_{t-1},y_t,x;\lambda) = \lambda^{1}f(y_{t-1},y_t) + \lambda^{2}f(y_t,x)
\end{equation}
where $\lambda=\left(\lambda^1,\lambda^2\right)$ are the model parameters to be estimated during training and $f(.)$ are transition and emission feature functions that produce boolean values. The term $Z(x)$ is the partition function and normalizes the conditional probability.

During model training, we seek to maximize the conditional log likelihood $\mathcal{L}$, defined as:
\begin{align}
\mathcal{L}(\lambda) &= \log p(y|x), \\
\mathcal{L}(\lambda) &= \sum^{T}\left(\lambda^{1}f(y_{t-1},y_t) + \lambda^{2}f(y_t,x)\right)-\log(Z(x)).
\end{align}

Since $\mathcal{L}$ is a convex function we apply a quasi-Newton method for estimation of model parameters $\lambda$ such as the L-BFGS optimization algorithm.  
The partition function considers a summation over all possible values of $x$ and $y$. 
We calculate the value of $Z(x)$ using the belief propagation (sum--product) algorithm  
which recursively calculates the passing of messages over all elements in the tree. 
In the case of linear chain graphical models, belief propagation provides an exact solution for the calculation of $Z(x)$, given by $Z(x)=\sum_{y_T}\alpha_T$, where $\alpha_t$ are the messages propagating forward in the algorithm.

\subsection[Dynamically Weighted Conditional Random Fields (dWCRF)]{\texorpdfstring{Dynamically Weighted Conditional Random Fields\\ (dWCRF)}{Dynamically Weighted Conditional Random Fields (dWCRF)}} 
\label{ssec:WCRF}

In this section we detail our dynamically weighted CRF (dWCRF) approach for structured predictions. 
The main motivation for the implementation of a weighted approach is to address the negative effects of imbalanced data on learning. 
In addition, classification performance measurements such as accuracy (1-error) is not a suitable metric as results are biased towards the majority class. 
Hence, we require the minimization of false positives $(FP)$  or false alarms; and false negatives $(FN)$ or missed classifications. 
Intuitively, this means increasing both true positives $(TP)$ and true negatives $(TN)$ of all participating classes. 
We use the expected \fsc (or F-measure) \cite{Dimitroff2014}, as an optimization metric for our model as it considers $FN$ and $FP$ in its definition. 

In this context, we introduce a cost term in our classifier objective function to give a higher cost to errors in the minority classes to reduce the effects of imbalance on the classifier. 
We consider the weighted log-likelihood function
\begin{align}
\label{eq:LL}
\mathcal{L}(\lambda,w) &= \sum_{t=1}^{T}w_t\log p(y_t|x_t,\lambda) 
\end{align}
where $w_t$ is a scalar weight for each element of the training sequence of length $T$ (weight vector represented by $[w_t]_{t=1}^T$). 
In \cite{Dimitroff2014}, Dimitroff et al.\ demonstrated 
using the Pareto optimality concept that there exists a set of weights $w_{\text{F}_{\beta}}$ for which $\lambda_F$, the parameter that optimizes the expected F$_{\beta}$-score, coincides with weighted maximum likelihood optimization parameter $\lambda_{ML}^w$. 
More information on Pareto efficiency can be  found in \cite{Dimitroff2014,Ehrgott:2005}.
However, the approach followed in \cite{Dimitroff2014} was for a binary Maximum Entropy classifier. 
In this article, we extend this previous work to multiclass classification using dWCRF; 
we consider the expected overall \textit{F}$_{\beta}$-score ($\bar{F}$), given by the mean expected \textit{F}$_{\beta}$-score over all participating classes as
\begin{align}
\label{eq:F_prom}
\bar{F}= \frac{1}{K}\sum_{k=1}^{K}\left(\frac{(1+\beta^2)\text{Precision}_k.\text{Recall}_k}{\beta^2\cdot\text{Precision}_k+\text{Recall}_k}\right) 
\end{align}
where $\beta \in \mathbb{R}$ is non-negative and balances the contributions from precision and recall. 
Henceforth, for simplicity, we consider $\beta=1$, where precision and recall have the same influence; \ie\ the harmonic mean of both precision and recall. 
Let us assume $\widehat{\lambda}$ to be the maximizer of $\bar{F}$, 
and considering that $TP_k = \sum_{i:y_i=k}p(y_i=k|x_i,\lambda)$ and $FP_k =  \sum_{i:y_i\neq k}p(y_i=k|x_i,\lambda)$; expanding and operating in  (\ref{eq:F_prom}), can be rewritten as 
\begin{align}
\label{eq:F_lambda}
\bar{F}(\lambda)=\frac{2}{K}\left[\frac{TP_1}{TP_1+T_1+FP_1}+ \cdots + \frac{TP_K}{TP_K+T_K+FP_K}  \right]
\end{align}
where 
$T_k$ are the number of elements of class $k$ in the training sequence, \ie $\sum_k(T_k)=T$. 
From (\ref{eq:F_lambda}), we want to show that $\widehat{\lambda}$ is an element of the Pareto optimal set of the multicriteria optimization problem (MOP) 
\begin{align}
\label{eq:mop1}
\max_{\lambda}\{TP_1, TP_2, \cdots , TP_K\}.
\end{align}

We do not consider the $FP$ term from (\ref{eq:F_lambda}) 
as we are interested in maximizing $TP$s and reducing $FP$s; moreover, increasing $TP_{\{1\cdots K\}\setminus u}$ (set of all $TPs$ except that for class $u$) will reduce $FP_{k=u}$. 
If we consider that $\widehat{\lambda}$ is not Pareto efficient in the MOP in (\ref{eq:mop1}), then there is a $\lambda_0$ such that $(TP_1(\lambda_0),\cdots ,TP_K(\lambda_0))$ dominates $(TP_1(\widehat{\lambda}),\cdots ,TP_K(\widehat{\lambda}))$; \ie at least one of the objectives is improved by $\lambda_0$ compared to that of $\widehat{\lambda}$. 
Since the expression in (\ref{eq:F_lambda}) increases as $TP_k$ increases, implying that $\bar{F}(\lambda_0)>\bar{F}(\widehat{\lambda})$; this contradicts the initial assumption that $\widehat{\lambda}$ maximizes $\bar{F}$. 

We can also observe that the Pareto optimal set of (\ref{eq:mop1}) is contained in that of the MOP 
\begin{align}
\label{eq:mop2}
\max_{\lambda}\{p(y_1|x_1,\lambda), p(y_2|x_2,\lambda) \cdots , p(y_T|x_T,\lambda)\}
\end{align}
this is because if we assume a $\lambda$ that is Pareto optimal for (\ref{eq:mop1}) but not for (\ref{eq:mop2}), then we have a $\lambda_0$ that improves at least one of the objectives in (\ref{eq:mop2}) without decreasing the others. 
This means the \textit{K}-tuple $(TP_1(\lambda_0),\cdots, TP_K(\lambda_0))$ dominates $(TP_1(\lambda),\cdots, TP_K(\lambda))$, which contradicts the assumption that $\lambda$ is Pareto optimal for (\ref{eq:mop1}). 
Hence the $\bar{F}$ optimizer $\widehat{\lambda}$ is also Pareto optimal for (\ref{eq:mop2}) and therefore $\widehat{\lambda}$ is Pareto optimal for the MOP 
\begin{align}
\label{eq:mop3}
\max_{\lambda}\{\log p(y_1|x_1,\lambda), \cdots ,\log p(y_T|x_T,\lambda)\}
\end{align}
given that $\log(.)$ is a strictly increasing function. 
The Pareto optimal set of (\ref{eq:mop3}) can be obtained by maximizing non-negative linear combinations of its objectives \cite{Ehrgott:2005}.  
This means there is a set of weights ($w_1,\cdots ,w_T$) such that
\begin{align}
\label{eq:lambda_mop}
\widehat{\lambda_{\bar{F}}} &= \argmax_{\lambda}\left( \sum_{t=1}^{T}w_t\log p(y_t|x_t,\lambda) \right) \nonumber\\ 
&= \argmax_{\lambda} (\mathcal{L}(\lambda,w) ) 
\end{align}
where the rightmost expression corresponds to the weighted log-likelihood as expressed in (\ref{eq:LL}). 
Our work above expands the proof in \cite{Dimitroff2014} for binary classification to multiclass classification.

\subsection{Weights Estimation}
\label{sssec:weights}
Now we are interested in computing the set of weights $w$ in dWCRF that maximizes the function $\bar{F}$. 
We use the previous result in (\ref{eq:lambda_mop}), which indicates that the objective functions $\bar{F}$ and weighted log-likelihood have gradients equal to zero at the optimal $\widehat{\lambda}$. 
We have the gradient of the function $\bar{F}$
\begin{dmath}
    \label{eq:grad1}
    \nabla_{\lambda}\{\bar{F}(\widehat{\lambda})\} = \sum_{t:y_t=1}\partial_{TP_1}\bar{F}(\widehat{\lambda})\nabla_{\lambda}p(y_t|x_t,\widehat{\lambda}) + \cdots  +  \sum_{t:y_t=K}\partial_{TP_K}\bar{F}(\widehat{\lambda})\nabla_{\lambda}p(y_t|x_t,\widehat{\lambda}) 
\end{dmath}
and the gradient of the log-likelihood function:
\begin{equation}
\label{eq:grad2}
\nabla_{\lambda} \mathcal{L}(\widehat{\lambda}) = \sum_{t=1}^{T}\frac{w_t}{p(y_t|x_t,\widehat{\lambda})} \nabla_{\lambda}p(y_t|x_t,\widehat{\lambda})
\end{equation}
where  $\nabla_{\lambda}\{\bar{F}(\widehat{\lambda})\}=\nabla_{\lambda} \mathcal{L}(\widehat{\lambda})=0$ at the optimal parameter $\widehat{\lambda}$.

Considering the expression in (\ref{eq:grad1}) we can obtain the partial derivative:
\begin{dmath}
    \label{eq:partDev}
    \frac{\partial F(\lambda)}{\partial TP_k} = 
    q\left[ \frac{d}{d TP_k}\left( \frac{TP_1}{TP_1+N_1+FP_1} \right) \right. +  
    \cdots  +
    \left. \frac{d}{dTP_k}\left( \frac{TP_K}{TP_K+N_K+FP_K} \right) \right]
\end{dmath}
where $q=2/K$ and given that previously we have considered $FP_k$ to be a function of all $TP$s other than $k$, (\ref{eq:partDev}) can be expressed as:
\begin{dmath}
\label{eq:partDev2}
\frac{\partial F(\lambda)}{\partial TP_k} = 
q\frac{N_k+FP_k}{(TP_k+N_k+FP_k)^2} + 
q\sum_{j:1\cdots K \setminus k} \frac{-TP_j\frac{d FP_j}{d TP_k}}{(TP_j+N_j+FP_j)^2}
\end{dmath}
we consider that the derivative term in (\ref{eq:partDev2}) are close to zero at the optimal  $(\lambda\rightarrow \widehat{\lambda})$ and that the derivative is much smaller than its quadratic denominator term; hence we eliminate the summation term from (\ref{eq:partDev2}), resulting in 
\begin{equation}
\label{eq:partDev3}
\frac{\partial F(\widehat{\lambda})}{\partial TP_k} \simeq 
q \frac{N_k+FP_k}{(TP_k+N_k+FP_k)^2}.
\end{equation}

The resulting weights for optimizing the weighted maximum likelihood and the corresponding expected overall \fsc  can now be defined as:
\begin{equation}
\label{eq:weights_cases1}
w_i =
\begin{cases}
p(y_i=1|x_i,\widehat{\lambda})\partial_{TP_1}\bar{F}(\widehat{\lambda}) & \text{if } y_i = 1 \\
\qquad \cdots  &					\\
p(y_i=K|x_i,\widehat{\lambda})\partial_{TP_K}\bar{F}(\widehat{\lambda}) & \text{if } y_i = K.
\end{cases}
\end{equation}

We also consider a parameter $\tau>0$ that corresponds to the number of times $\mathcal{L}$ is computed during the optimization process. 
We use this parameter to apply $\lambda$ considerations for 
(\ref{eq:partDev3}), by executing first homogeneous weights as in linear chain CRF. 
Hence the resulting weights have the form:

\begin{equation}
w_{i,\tau} =
\begin{cases}
q & \text{if the number interations} < \tau  \\
w_i \text{, as in (\ref{eq:weights_cases1})} & \text{if the number interations} \geq \tau. 
\end{cases}
\end{equation}

\subsection{Real Time Inference}
\label{ssec:infere}

Usually, class inference process for CRF is performed for complete sequences of data where methods as the forward-backwards or Viterbi algorithms are applied to the test segment and complete sequence of labels is returned \cite{Sutton2012}. 
In a previous study, we have underscored the importance of real time prediction of activities \cite{ShinmotoTorres2013} where we applied the belief propagation method to obtain the marginal probabilities of the last received observation; we use the current sensor observation and the information from the last inference made on the previous observation. 
This is given by the form: 
\begin{dmath}
m(y_t) = \frac{1}{Z_t}\left( \exp\left( \phi \left( y_t,x \right) \right)  {\sum_{y_{t-1}}\left(\exp \left( \phi \left( y_{t-1},y_t \right) \right)m\left(y_{t-1}\right) \right)} \right)
\end{dmath}
where $m(y_t)$ is the marginal probability corresponding to the $t^{th}$ observation $x_t$ and $Z_t$ corresponds to the normalizing term so the marginals at a given time $t$ sum to unity and prevents the occurrence of floating point underflow. 
The assigned label corresponds to the activity that has the highest marginal probability.

\section{Experimental Studies}
\label{sec:experim}
This section presents the experimental framework and corresponding results. 
We evaluate our dWCRF method using datasets from two different human activity recognition approaches:
\begin{inparaenum}[i)]
\item using battery powered body worn sensors (BPBW); and \item batteryless body worn sensors (BLBW). 
\end{inparaenum}

\subsection{Problem Description} 
\label{ssec:seqdata}
Both scenarios, BPBW and BLBW, consider sequential data from the sensors; these are time series $X=\{x_t\}_{t=1}^{T}$, where $x_t\in \mathbb{R}^d$, 
and associated with a sequence of activity labels $Y = \{y_t\}_{t=1}^T$, where $y_t\in \mathcal{Y}=\{1\cdots K\}$, and $K$ is the number of activity labels to predict. 
In sequential data problems, we assume that sequences are i.i.d. from each other; that is, given the set of $M$ training training labeled sequences  $\mathcal{D}=\{(X_m,Y_m)\}_{m=1}^M$, variables in $(X_i,Y_i)$ are independent of those in $(X_j,Y_j)$ for $i\neq j$. 
However, dependency relationships between variables in a sequence cannot be assumed. 
Hence, given a testing sequence $\mathcal{T}=\{(X,Y)\}$, we are interested in predicting individual class labels $\hat{y}_t$ for every individual observation $x_t$ using our trained dWCRF model.

\subsection{Statistical Analysis}
\label{ssec:stat}
In this study, we determine class specific performance measurements: true positives ($TP$) are the correctly predicted activity labels. 
False positives ($FP$) are those predicted labels that are misclassified, thus do not match with the ground truth. 
False negatives ($FN$) correspond to those ground truth classes that were missed. 
True negatives ($TN$) are those non-target (not intended) classes that were correctly identified by the system. 

In addition, we evaluate the performance of each class $k$ using the harmonic mean of Precision ($Pr$) and Recall ($Re$):
\begin{align}
{\precis}_k (Pr) &= TP_k/\left( TP_k +FP_k\right) \\
{\recall}_k (Re) &= TP_k/\left( TP_k+FN_k \right) \\
\thickmuskip=0.5\thickmuskip
{\fscore}_k =& {\frac{2 {\times} Pr_k {\times} Re_k}{Pr_k + Re_k}} = \left( \dfrac{1}{2{\times} Pr_k}  + \dfrac{1}{2{\times} Re_k} \right)^{-1}.
\end{align}
In the case of overall performance, we consider the average of the class-specific performance metrics \ie $\fscore_{Overall}= \sum_{k}\fscore_k / |k|$. 

Note we do not evaluate metrics depending on true negatives ($TN$) such as specificity \cite{Najafi2003ambulatory,Godfrey2011,Shinmoto:6548154}  
as specificity does not appropriately reflect the performance of the minority class. 
This is because $TN$ considers all activities other than the target activity. 
For example $TN$ of a minority class includes the $TP$ of the majority class, producing high specificity values,  giving an over optimistic measurement of performance.

We are interested in comparing the \fsc results of our classifier with other classifiers. 
Therefore, we compare the significance between two classification results using a two-tailed independent t-test. A p-value (\textit{p}) \textless 0.05 is considered statistically significant.

Evaluation of these metrics in the case of the BLBW datasets was performed using a 10-fold cross validation procedure, where each fold considered complete sequences of activities (a trial) of different people. 
We considered 6 folds for training 2 folds each for testing and validation. 
In the case of the BPBW datasets, these were evaluated using a 4-fold cross validation for each dataset, due to the reduced number of trials per dataset; we use two folds for training and one for testing and validation respectively.

\subsection{Batteryless Body Worn Sensor Datasets (BLBW)}

These datasets were obtained in the context of a larger project by our research group directed at the ambulatory monitoring of hospitalized older patients to prevent falls \cite{Visvanathan_6347326}. 

We evaluate three case studies based on motion information from trialled healthy and hospitalized participants using the battery-less body worn sensor \cite{Kaufmann2013}, shown in Figure~\ref{Fig:Multifigu}(a). 
Trial participants were requested to perform a series of broadly scripted ADLs which included:  
\begin{inparaenum}[i)]
    \item Sitting on bed;
    \item Sitting on chair; 
    \item Lying on the bed; and
    \item Walking to the bed, chair or door.
\end{inparaenum} 
These represent the most likely activities performed in a hospital environment by older patients. 
A researcher, present during the trials, annotated the labels directly into the middleware for reference as ground truth.

We consider that posture transitions such as sit to stand and stand to sit are integrated into the ambulation or sitting movements as data collected during posture transitions is scarce as the movements are of short duration \cite{Shinmoto:6548154}. 
For example, we consider a participant starts ambulating as soon as the body is not in contact with the bed or chair. 
Ambulation, in this case, also includes standing and any movement the participant performs while walking around the room. 

Therefore, given the limitations imposed by the physical space and movement of our target demographics, the BLBW datasets consider four classes $(K=4)$ to distinguish whether a person is in or has exited a resting posture.  
These classes are: 
\begin{inparaenum} [i)]
    \item Sit-on-bed;
    \item Sit-on-chair; 
    \item Lying; and
    \item Ambulating; 
\end{inparaenum} 
where class Ambulating includes all other movements associated with sitting on chair, bed or lying. 
Details regarding the sensor platform and the case studies are explained below.

\subsubsection{Sensor Platform}
\label{ssec:sensor}
\begin{figure}
    \centering
    \renewcommand{\tabularxcolumn}[1]{m{#1}}
    \addtolength{\tabcolsep}{-3pt} 
    \begin{tabularx}{1\columnwidth}{@{}XX@{}} 
        \begin{tabular}{c}
            \phantomcaption
            \begin{subfigure}[]{0.49\columnwidth}
                \centering
                \includegraphics[width=3.5cm]{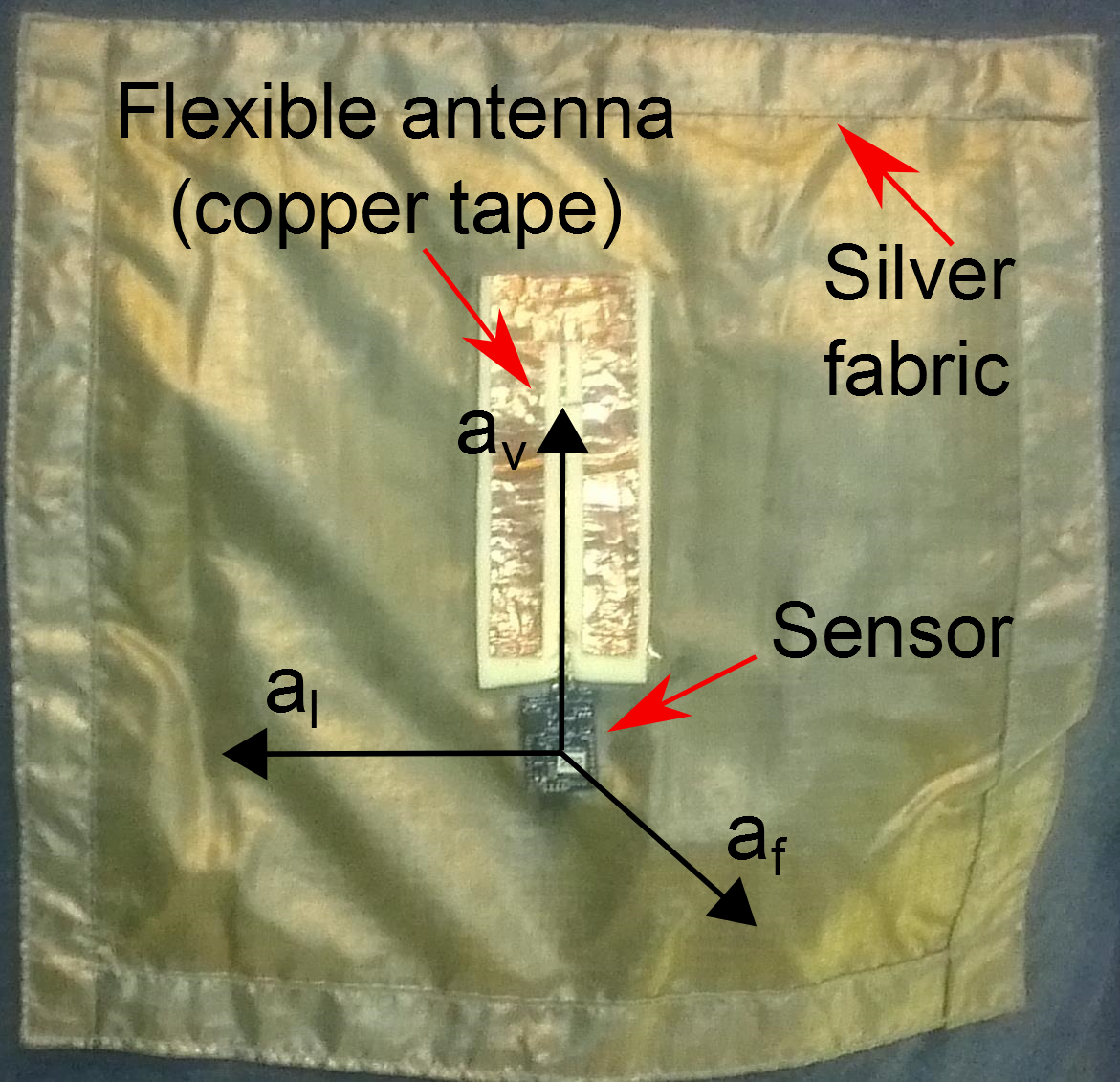}
                \caption{}
                \label{Fig:wisp2}
            \end{subfigure}
            \\
            \begin{subfigure}[]{0.49\columnwidth}
                \centering
                \includegraphics[width=3.8cm]{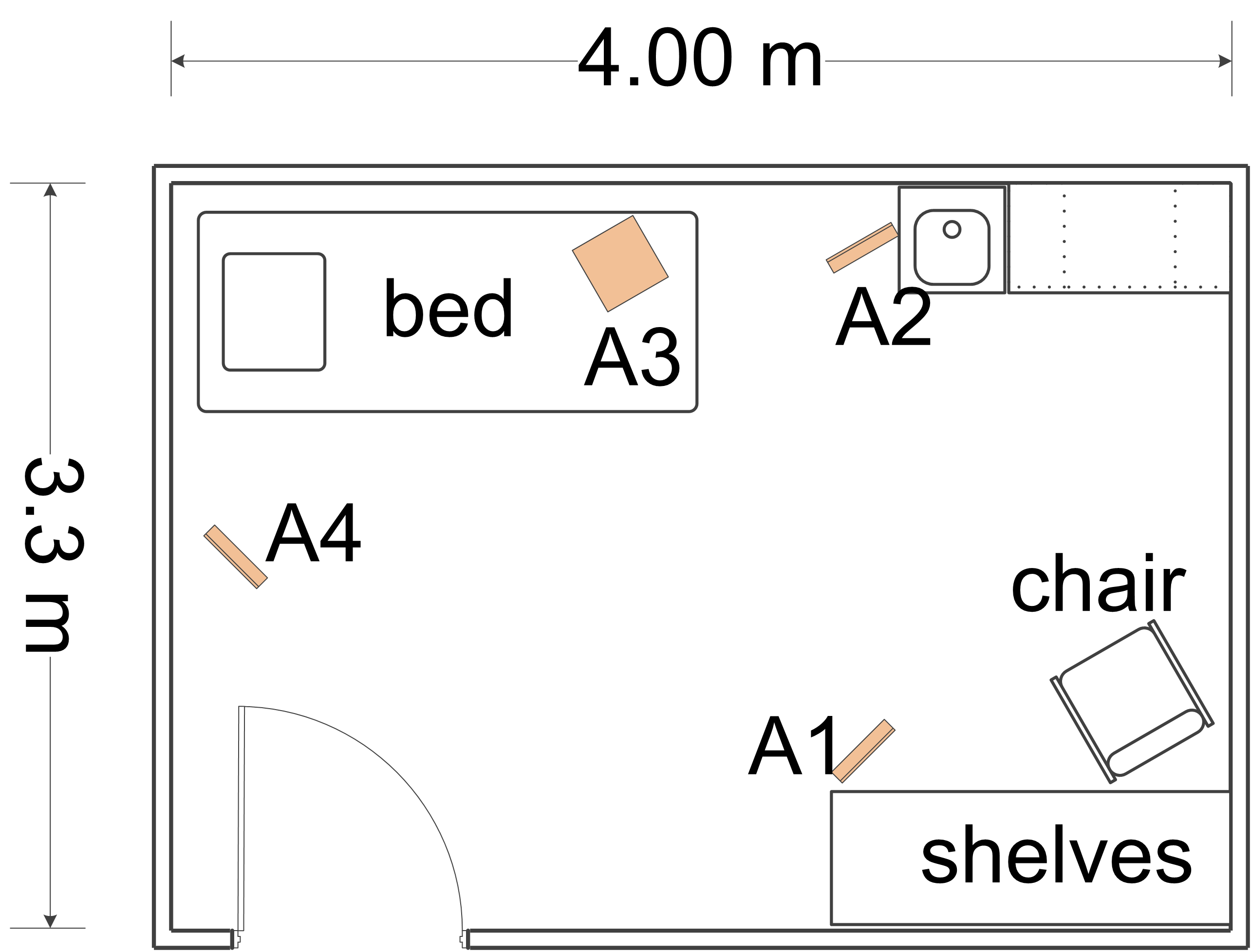}
                \caption{}
                \label{Fig:Datasets1}
            \end{subfigure} \\
            \begin{subfigure}[]{0.49\columnwidth}
                \centering
                \includegraphics[width=3.8cm]{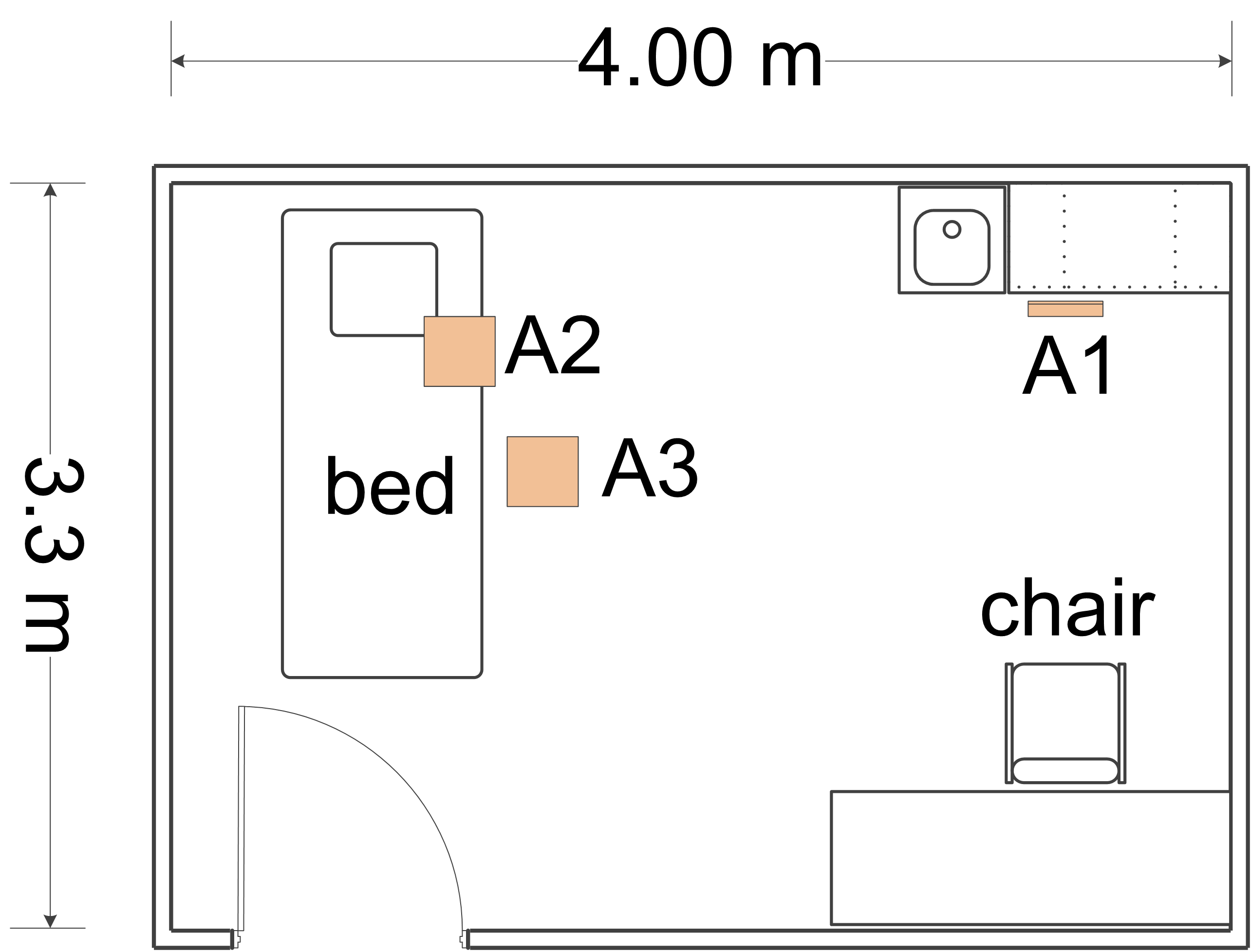}
                \caption{}
                \label{Fig:Datasets2}
            \end{subfigure} \\
            \begin{subfigure}[]{0.49\columnwidth}
                \centering
                \includegraphics[width=3.8cm]{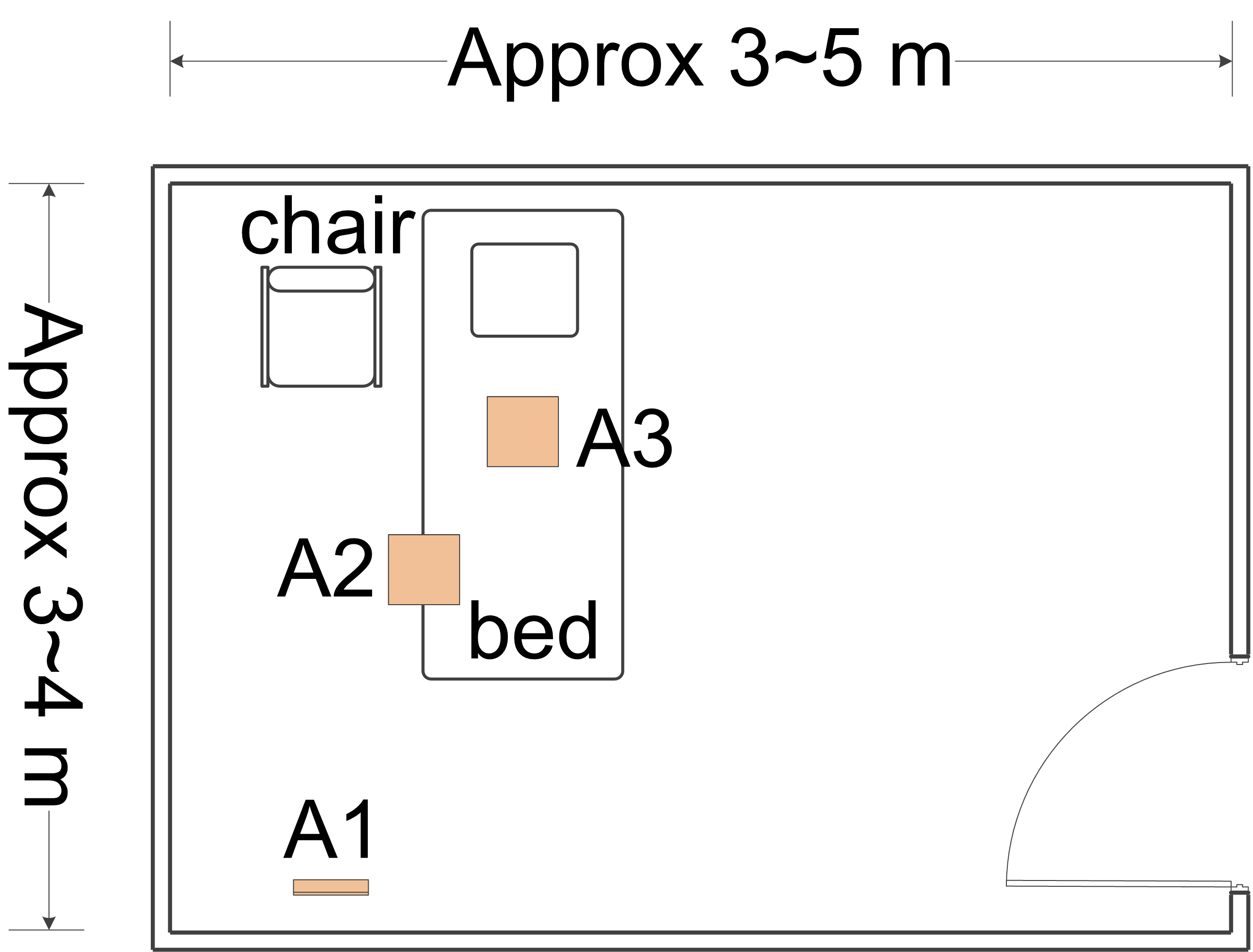}
                \caption{}
                \label{Fig:Datasets3}
            \end{subfigure}
        \end{tabular} %
        &
        \begin{tabular}{c}
            \begin{subfigure}[]{0.49\columnwidth}
                \centering
                \includegraphics[width=4.0cm]{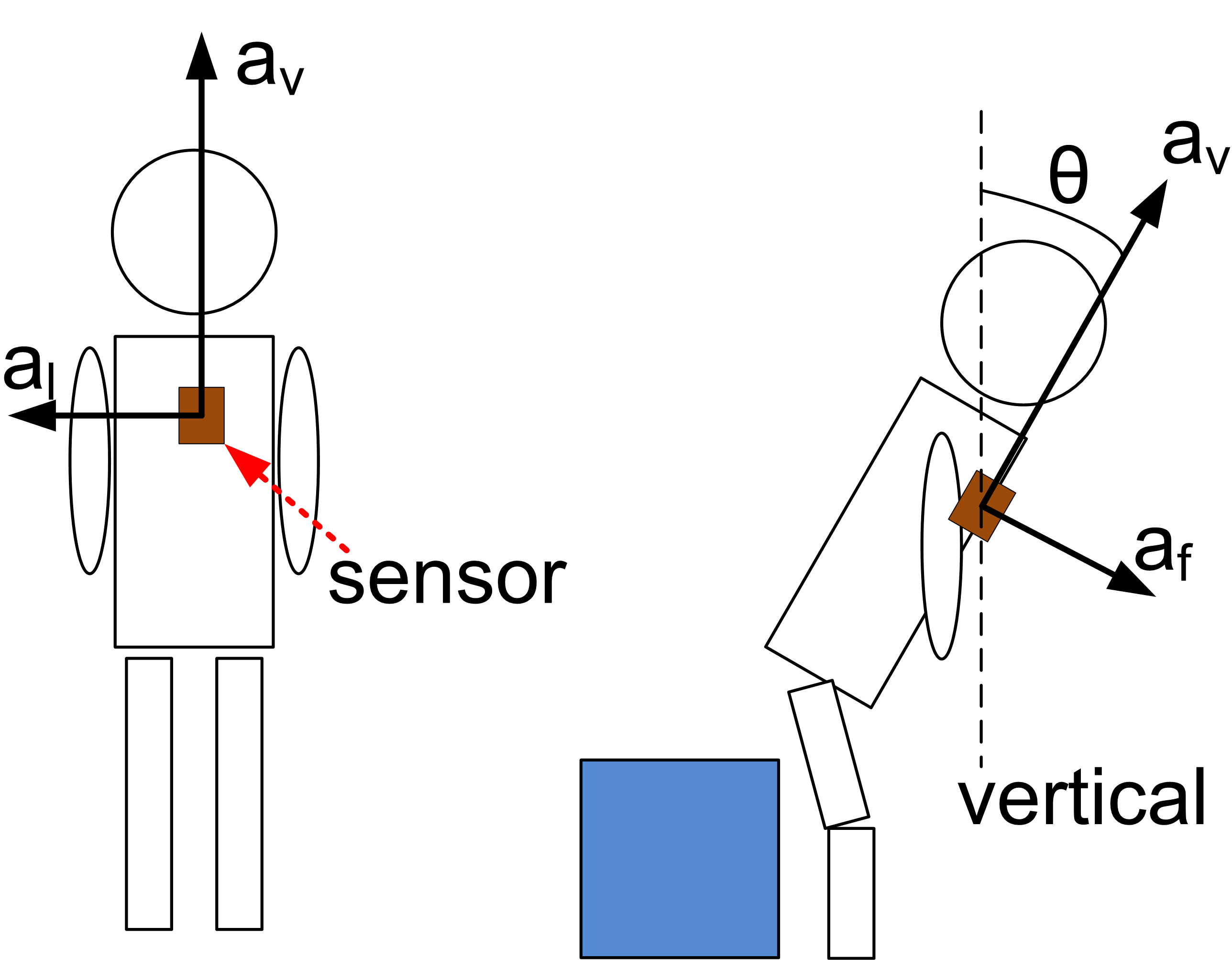}
                \caption{}
                \label{Fig:human}
            \end{subfigure} \\\\
            \begin{subfigure}[]{0.49\columnwidth}
                \centering
                \includegraphics[width=4.0cm]{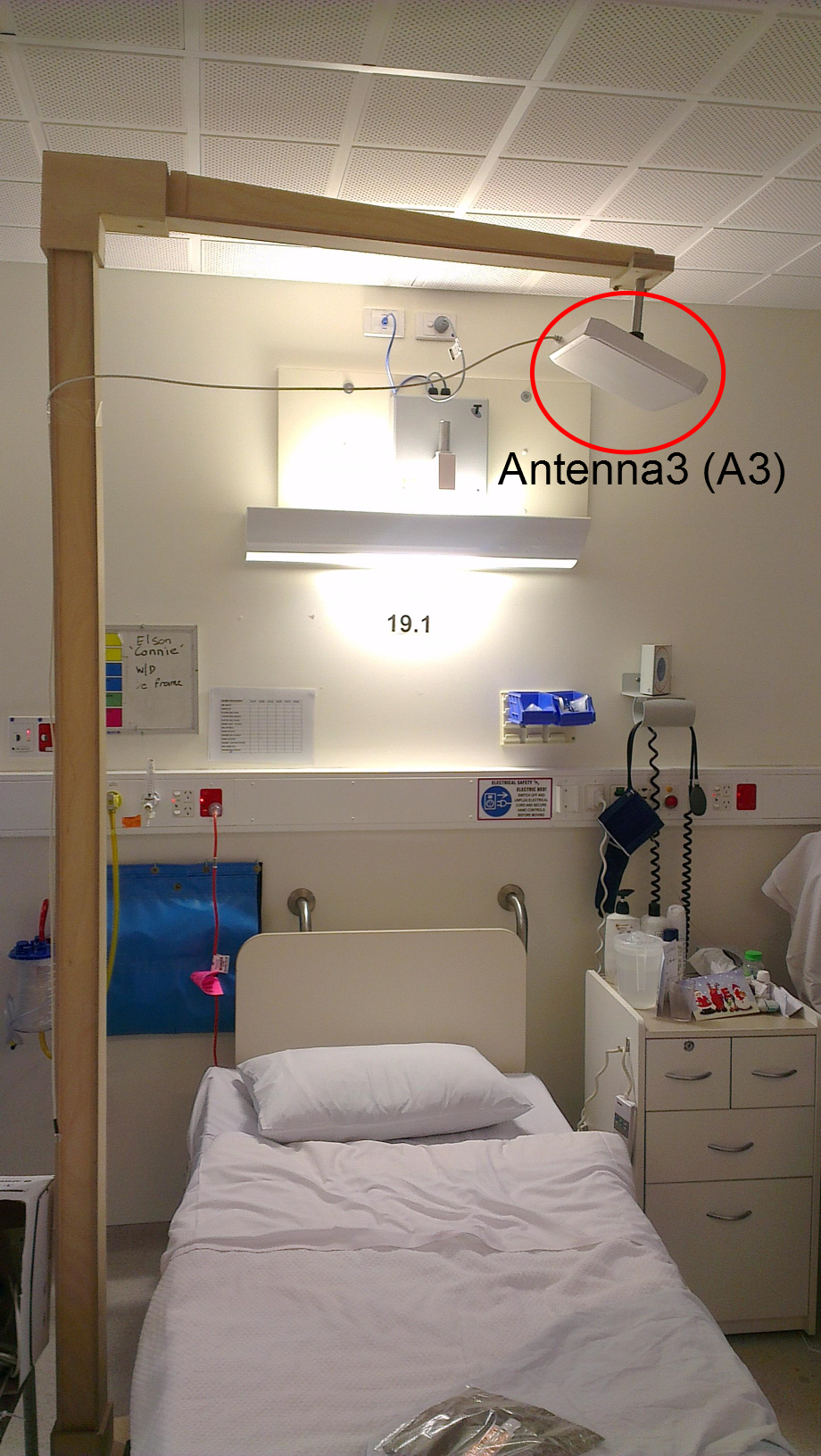}
                \caption{}
                \label{Fig:Room3Conf}
            \end{subfigure}
        \end{tabular} %
    \end{tabularx}
    \addtolength{\tabcolsep}{3pt} 
    \caption{(a): Wearable sensor (\wisp) and acceleration vector components. (b)--(d): Physical deployments of an RFID antenna infrastructure (antennas A1, A2, A3, A4) for collecting three datasets: (b): healthy older people (Room1); (c): healthy older people (Room2) and (d): hospitalized older people (Room3). (e): Front and lateral view of sensor axes for a participant standing and sitting/standing with respect to vertical.  (f): Bed overview of a Room3 case, showing antenna A3 and supporting frame.}
    \label{Fig:Multifigu}
\end{figure}

The participants wore a flexible Wearable Wireless Identification and Sensing Platform (\wisp) device, developed by our team \cite{Kaufmann2013}, over a garment on top of the sternum. 
The \wisp, see Figure~\ref{Fig:Multifigu}(a), and based on \cite{Sample2008design},  
encases a tri-axial accelerometer (ADXL330) and a 16 bit microcontroller (MSP430F2132). 
The \wisp\ is part of an emerging class of batteryless sensors. 
In particular, the \wisp\ harvests its energy using the electromagnetic field illuminating the tag from RFID antennas, which also collect the \wisp\ sensor data. 
The main motivation for using this passive (batteryless) device compared to using other battery powered sensors are twofold: 
\begin{inparaenum}[i)]
    \item the device requires no maintenance as it is battery free, lightweight, inexpensive and easy to replace; and 
    \item frail older people, especially those with conditions such as delirium or dementia, require easy-to-use equipment \cite{Topo20095}, and our proposed sensors' development objective is to be inconspicuous to the user \ie concealed in the clothes.
\end{inparaenum} 

We collect tri-axial acceleration signals and the received signal strength indicator (RSSI) from the sensor signals. 
RSSI is used as a measure of relative distance to the antenna receiving a sensor observation, specially over a short distance as in our case studies. 
Moreover, because the device is passive, sensor observations are not regularly collected in time, and thus 
increasing the complexity of the problem (see Figure~\ref{Fig:DataProb}).

\subsubsection{Case Studies}
\paragraph{Case Study 1}
Fourteen healthy older volunteers, with average age of 74.6 $\pm$ 4.9 years old, completed around five trials each, based on their ability and level of fatigue. 
Participants were allocated into two different room configurations: Room1 and Room2, each constituting a dataset with  4 and 3 antennas deployments, shown in Figure~\ref{Fig:Multifigu}(b) and (c), respectively.

\paragraph{Case Study 2}
Twenty five hospitalized patients, with average age of 84.4 $\pm$ 5.3 years old, 
performed a short sequence of ADLs due to the frailty of the participants. 
The patients were trialled in their respective rooms, constituting dataset Room3. 
In this hospital room configuration, see Figure~\ref{Fig:Multifigu}(d), 
the displayed measurements are approximate due to differences between rooms (single or double bed), and the bed and the chair were always next to each other.

\paragraph{Case Study 3}
This case study investigates the performance of our method under the conditions of reduced training data; we use one dataset, Room1 from Case~Study~1, where data sequences are extracted to simulate datasets of increasingly reduced number of sensor observations. 

\subsubsection{Class Imbalance in Datasets}
\label{ssec:classimb}
\begin{figure}[tb] 
    \centering
    \includegraphics[width=0.6\linewidth]{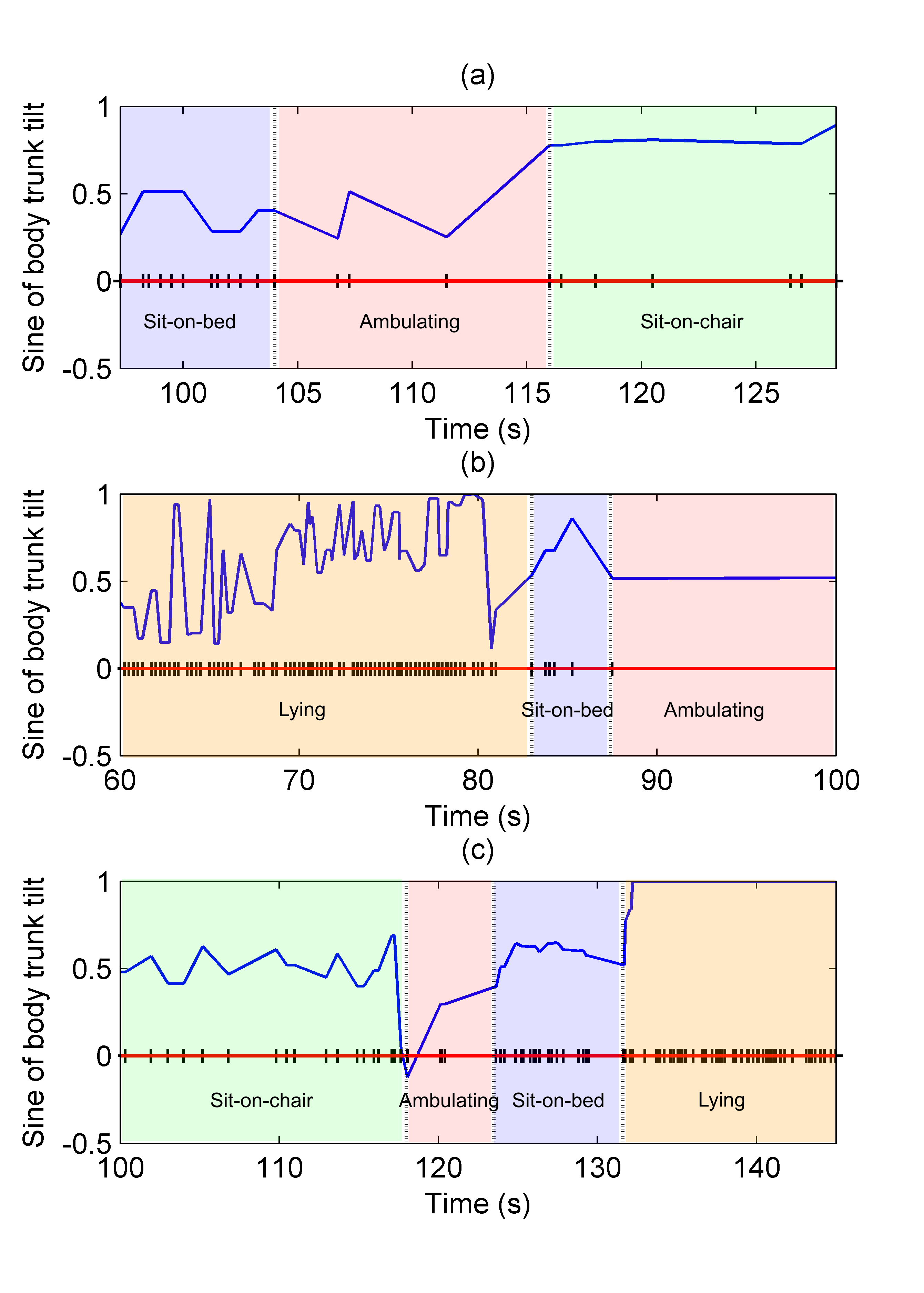} 
    \caption{Raw data corresponding to body tilting with respect to the vertical for the three datasets (a) Room1, (b) Room2 and (c) Room3; where black vertical marks represent sensor readings and classes are: Sit-on-bed (blue), Ambulating (red), Sit-on-chair (green) and Lying (orange).
        } 
    \label{Fig:DataProb}
\end{figure}

\begin{figure}[tb]
    \centering
    \includegraphics[]{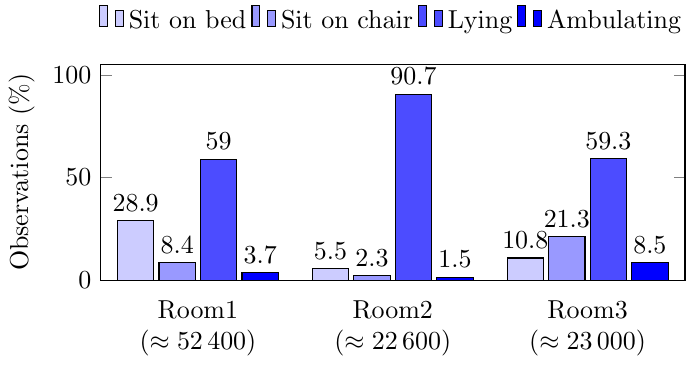}
    \caption{Overview of the data imbalance in the three datasets: Healthy older people Room1; Healthy older people Room2; and Hospitalized older people Room3. In all datasets the majority class is Lying and minority class is Ambulating. In brackets are shown the approximate number of sensor observations per dataset.}
    \label{Fig:Imbalance}
\end{figure}

Two main sources of imbalance affect our datasets. 
The first is the duration of different activities. 
This is expected as, for example, lying on bed is of longer duration than ambulating. 
The second source of imbalance is due to the passive nature of the \wisp\ sensor; this affects the powering of the device and the regularity of sensor readings. 
Sensor positioning and proximity to RFID antennas, posture of the participant (causing occlusion) can also affect the powering of the sensor; moreover, these conditions can change from person to person and room to room.

From the room settings, we can see that Room1 
intends to collect sensor observations from the complete room, whereas Room2 and Room3 are focused on obtaining data from specific areas around the bed and chair while saving on hardware infrastructure.  
In Room3, the small dimensions of the path between bed and chair cause ambulation time to be minimal.

An illustration of sensor observations and the resulting data imbalance is shown in Figure~\ref{Fig:DataProb}. 
Data acquired from Room1, see Figure~\ref{Fig:DataProb}(a), indicates that sensor inter-reading times when the participant is sitting on bed range from 0.2\,s to 1.3\,s, from 0.5\,s to 4.2\,s when the person is ambulating and 0.5\,s to 6\,s when sitting on a chair. 
In addition, the first observation corresponding to Ambulating and Sit-on-chair are received after 0.7\,s and 4.5\,s respectively.

In the case of Room2 and Room3, shown in Figure~\ref{Fig:DataProb}(b) and (c), sensor observations from a person lying in bed (transparent-orange background) are more frequently collected due to the location of the antennas when compared to a person sitting on bed (transparent-blue background) or ambulating (transparent-red background).

Class imbalance of the datasets are shown in Figure~\ref{Fig:Imbalance}. 
Datasets Room1 and Room3 show similar imbalance where there is a dominant class (Lying), a second dominant class (Sit-on-bed with 28.9\% and Sit-on-chair  with 21.3\% respectively) and the minority class Ambulating has the lowest proportion in both datasets.
However, Room1, has more than double the number of sensor observation of Room3 (see Figure~\ref{Fig:Imbalance}); and the minority class (Ambulating) for both datasets has the same number of observations. 
Dataset Room2 is almost dominated by one class (Lying) and the rest of the classes have decreasing values of participation with Ambulating the minority class (1.5\%). 
This dataset has almost the same amount of sensor readings as Room3;  
albeit Room3 having collected data from more participants than the other datasets.

\subsubsection{Feature Extraction}
\label{sssec:Feat}

From the three datasets we extract features representative of activities. 
We use a fixed time sliding window of duration of 4\,s (referred henceforth as segment) from which we extract instantaneous sensorial data corresponding to the last received observation and contextual information associated with observations in the segment. 
Moreover, we also use inter-segment information to further capture trend changes of the sensor signals.
We selected this window method and length as it is simple to deploy and performs as well as more complex windowing methods for feature extraction \cite{ShinmotoTorres2013}. 
Hence, the feature vector is composed of three different types of features:

\paragraph{Instantaneous Features}
These features are strictly obtained from the last received observation corresponding to the current performed activity, shown in Table~\ref{Table:Instant}, and the gender of the participant which is known. 

\begin{table}[tbp]
    \caption{Instantaneous features.}
    \label{Table:Instant}
    \begin{tabularx}{\columnwidth}{@{}>{\setlength\hsize{0.25\columnwidth}}X X@{}} 
        \multicolumn{1}{c}{Feature}   & \multicolumn{1}{c}{Description}     \\    \midrule
        $a_f$              & frontal acceleration                \\
        $a_v$              & vertical acceleration               \\
        $a_l$              & lateral acceleration                \\
        $\sin(\alpha)$      & sine of body tilting angle          \\
        $aID$              & receiving antenna identification    \\
        $RSSI$             & received signal strength indicator  \\
        $\Delta t$ & time difference with previous observation   \\
        $yaw$              & trunk yaw angle     \\
        $roll$             & trunk roll angle     \\
        $sex$              & participant's gender  \\      
        \bottomrule     
    \end{tabularx}
    \caption{Contextual information features.}
    \label{Table:Contextual}
    \begin{tabularx}{\columnwidth}{@{}>{\setlength\hsize{0.25\columnwidth}}X X@{}} 
        \multicolumn{1}{c}{Feature}   & \multicolumn{1}{c}{Description}    \\    \midrule
        $aSUM_{1..M}$    & number of events per antenna       \\
        $a_{max}RSSI$     & antenna collecting maximum received power   \\
        $a_{min}RSSI$     & antenna collecting minimum received power   \\
        $Vdisp$          & vertical displacement              \\
        $MI_{bed-chair}$ & mutual information of bed and chair areas       \\
        $r_{[fv,fl,vl]}$ & Pearson correlation coefficient for acceleration axes \\ 
        \bottomrule
    \end{tabularx}
    
    \caption{Inter-segment features.}
    \label{Table:Intersegment}
    \begin{tabularx}{\columnwidth}{@{}>{\setlength\hsize{0.25\columnwidth}}X X@{}} 
        \multicolumn{1}{c}{Feature}   & \multicolumn{1}{c}{Description}    \\    \midrule
        $\Delta Max[a_f,a_v,a_l]$   & Difference of acceleration maxima per axis  \\
        $\Delta Min[a_f,a_v,a_l]$   & Difference of acceleration minima per axis  \\
        $\Delta Med[a_f,a_v,a_l]$   & Difference of acceleration median per axis  \\
        $\Delta MaxRSSI_{1..M}$     & Difference of power maxima per antenna \\
        $\Delta MinRSSI_{1..M}$     & Difference of power minima per antenna \\
        $\Delta MedRSSI_{1..M}$     & Difference of power median per antenna   \\
        \bottomrule
    \end{tabularx}
\end{table}

We consider the body tilting angle $\alpha$ on the midsagittal plane towards the front or back of the participant from the vertical reference \cite{Najafi2003ambulatory}. 
The angle $\alpha$ is approximated from current acceleration values $\alpha \approx \arctan\left( \frac{a_f}{a_v}\right)$; however, we prefer $\sin(\alpha)$ as it is proportional to $\alpha$ and range limited to [-1,1]. 
The value of RSSI is of interest as a reference of relative proximity to a surrounding antenna. 
We also consider the time difference $(\Delta t)$  between sensor observations as in previous research \cite{Shinmoto:6548154}. 
Also included are the body rotation angles 
approximated from current acceleration readings: $yaw=\arctan(\frac{a_l}{a_f})$ and $roll=\arctan(\frac{a_l}{a_v})$, as they  carry information pertinent to body rotation movements.

\paragraph{Contextual Information Features}
These features, shown in Table~\ref{Table:Contextual}, are extracted from  each segment and provide general information of what is occurring during the segment duration, \ie complementary information to the instantaneous features in the temporal vicinity of the last received sensor observation \cite{Krishnan_2014}.

We include the basic contextual information of the number of events per antenna in the segment; 
we also include the identification of the antenna that registered the highest and lowest RSSI in the segment, which serve as a location marker as a participant is more likely to occupy an area near the antenna reporting higher RSSI during the segment duration. 
Other feature is the vertical displacement measured from acceleration readings in the vertical axis $(a_v)$ in the segment. 
The mutual Information between bed and chair areas considers the occurrences of consecutive observations from two antennas focused towards the chair and bed occurring in either directions as used in \cite{ShinmotoTorres2013}. 
We also consider the Pearson correlation coefficient of all combinations of the three acceleration components of all observations in the segment.

\paragraph{Inter-Segment Features}
These features, shown in Table~\ref{Table:Intersegment}, aim to capture information trend variations from consecutive segments and are useful as these variations are insensitive to noise in unfiltered raw sensor data.

We include the difference of the maxima, minima and median of the segments' acceleration readings in the three axes with respect to the participant: vertical, frontal and lateral. 
In addition, we are interested in the changes of RSSI, as an indicator of position shifting, given by the difference of the segments maxima, minima and median of the RSSI readings per antenna. 

\subsubsection{Parameter Selection}
We evaluate the performance of our dWCRF model together with linear chain CRF \cite{Lafferty2001}; a weighted CRF with fixed weight values (fWCRF), given by the inverse of the class distribution \cite{Lannoy_6036156}; and a cost parameter based CRF such as the softmax-margin model (C-CRF) \cite{Gimpel2010} on all three datasets, we use the L2 regularized model for each classifier of the form $\theta \lVert\lambda\rVert^2$. 
Regularization parameter $\theta$ was evaluated in the range $[10^{-4}, 10^{-1}]$. 
Parameter $\tau$ was chosen from the range limited by the lowest number of iterations for linear chain CRF. 

In addition, we also compare with multiclass SVM for linear and radial basis function (RBF) kernels (L-SVM and R-SVM respectively) and the weighted SVM for both classifiers (L-WSVM and R-WSVM) \cite{Cortes_SVM,Boser92atraining,Friedman:96}. 
All classifiers were trained using the extracted features described in Section \ref{sssec:Feat}. 
Selection of hyperparameters for the  SVM classifiers' regularization, $C$, and RBF kernel, $\gamma$, were evaluated using a grid search in the range $[2^{-5}, 2^5]$ for both parameters. 
SVM algorithms were evaluated using libSVM \cite{LibSVM_2011} toolbox in Matlab, which performs a one-vs-one approach for multiclass classification.
In the case of dWCRF, CRF, fWCRF, L-SVM and R-SVM, these are the only validated parameters; in all cases, the best set of parameters that produced the highest \fsc was chosen. 

The weight parameters for C-CRF, L-WSVM and R-WSVM, are found through cross-validation, evaluating the validation set. 
We note that for fWCRF, the weights are fixed and determined solely on the class distribution; and are not sought by any optimization method. 
In the case of C-CRF, the parameters are selected to optimize \fsc as in \cite{Gimpel2010a}; we applied an extensive grid search to obtain the optimal parameters in the value range $[0, 20]$. 
In the cases of weighted SVM, the algorithms require a weight per class, $K=4$ in our case, requiring a larger grid-search evaluation of the order of $N^4$ processes, where $N$ is the number of elements in the range to evaluate. 
Instead we use the covariance matrix adaptation evolution strategy (CMA-ES) \cite{hansen2006eda}, a widely used evolutionary optimization algorithm, to find the optimal set of per-class parameters for these classifiers.  
Given the stochastic nature of the initial parameter selection for the iterative CMA-ES process, we require  evaluating multiple starting values; in our case, we evaluated 350 random initial points uniformly distributed in the range $[0, 20]$ for each classifier.

Hyperparameter validation for the tested methods was performed on a cluster of Intel 8 core E5 series Xeon microprocessors due to the large number of processes to be performed.

\subsubsection{Results}
\label{BLBW:results}
\begin{figure}[tb]
    \centering
    \includegraphics[width=0.99\linewidth]{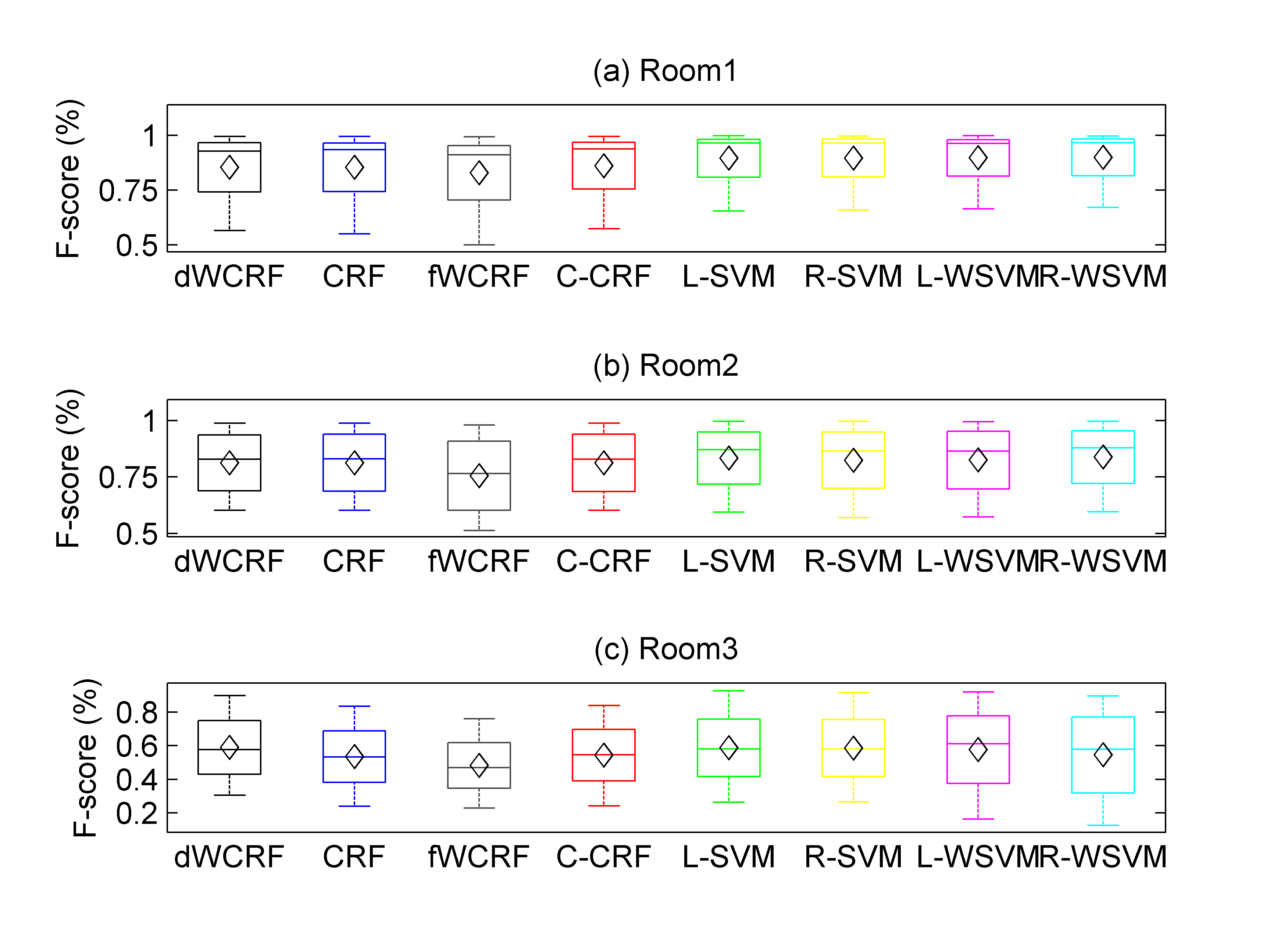}
    \caption{Overall \fsc performance of our three datasets (a): Room1,  (b): Room2 and (c): Room3  using our dWCRF, linear chain CRF, fixed weighted CRF (fWCRF), softmax margin CRF (C-CRF), linear SVM (L-SVM), RBF kernel SVM (R-SVM) and their corresponding weighted algorithms: L-WSVM and R-WSVM respectively. Results are shown as boxplots (averages shown as diamonds).}
    \label{Fig:Overall}
\end{figure}

First, we demonstrate the overall results corresponding to the datasets from Case~Study~1 (Room1 and Room2) and Case~Study~2 (Room3). 
These were obtained by averaging all participating classes' individual \fsc are shown in Figure \ref{Fig:Overall}. 
Maximum performance variations between methods' average \fsc is about $7\%$. 
For Room1, variation of our dWCRF with the best performing classifier, R-WSVM, is 4.6\% (85.4\% and 90\% respectively). 
For Room2, the maximum \fsc  variation is $\approx 8\%$ where the difference between our dWCRF (81.3\%) and best performing R-WSVM (83.8\%) is 2.5\%. 
In Room3, the maximum \fsc variation is $\approx 10\%$, where dWCRF is the best performing classifier (59.0\%).  
In addition, overall dWCRF results are not statistically significantly different to those of all other algorithms;  p-values for Room1 are $p>0.72$, for Room2 are $p>0.67$ and for Room3 are $p>0.53$. 
Therefore, dWCRF performs as well as other classifiers; particularly in Room3 where dWCRF mean \fsc was higher than other classifiers.

\begin{figure}

    \includegraphics[trim = 4.7cm 5.7cm 6.7cm 6cm,clip, width=0.85\linewidth]{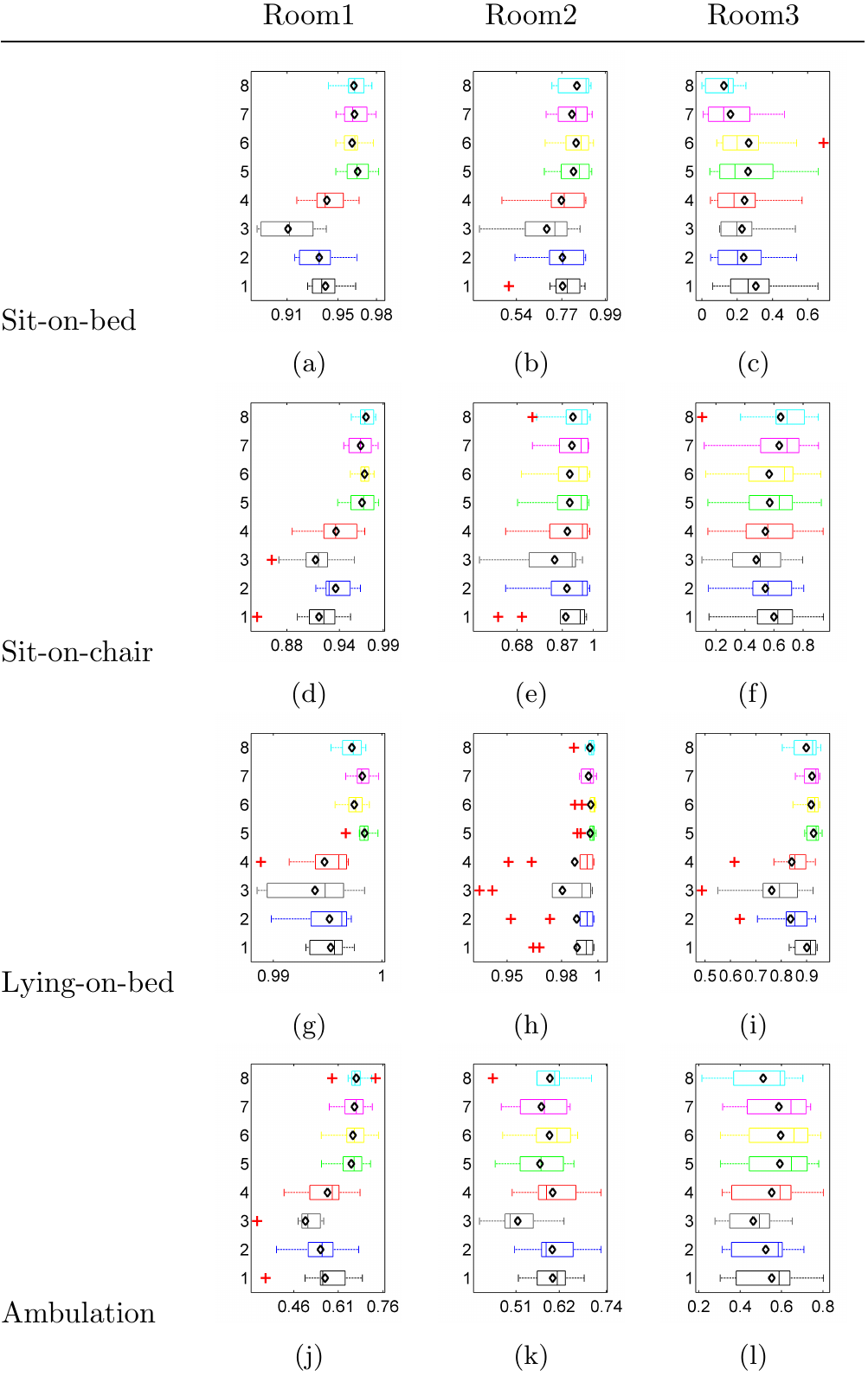}
    \caption{F-score performance of all classes and datasets. Results are shown as boxplots (averages shown as diamonds). Classifiers tested are: 1: dWCRF, 2: CRF, 3: fWCRF, 4: C-CRF, 5: L-SVM, 6: R-SVM, 7: L-WSVM, 8: R-WSVM.}
    \label{Fig:Individual-Rooms}
\end{figure}

In terms of individual classes from the datasets corresponding to Case~Study~1 and Case~Study~2, we observe differing behaviours. 
For Room1, see Figure~\ref{Fig:Individual-Rooms}(left column), SVM classifiers achieve statistically significantly better results than those of dWCRF ($p\leq0.015$). 
In particular, for the minority class (Ambulating), the difference in performance is $\approx 10\%$ in comparison to R-WSVM (see Figure~\ref{Fig:Individual-Rooms}(j)). 
For the rest of activities the differences are $<5\%$. 
In comparison to other CRF based algorithms, dWCRF has, in general, higher results in terms of mean \fsc with exception of class Sitting-on-chair, with difference of 2\% ($p>0.1$).

The results for Room2, shown in Figure~\ref{Fig:Individual-Rooms} (middle column), demonstrate  that all results are very close and no classifier is statistically significantly different from dWCRF ($p\geq 0.08$) with exception of Ambulation with fWCRF where dWCRF is better with statistical significance ($p=0.003$).  
For the minority class Ambulation, dWCRF mean performance is higher by 3\% compared to both L-SVM classifiers and 0.5\% compared to R-SVM classifiers. 
However, for the Sit-on-bed class, the R-WSVM result is higher by 7\%. 
Similarly to Room1, dWCRF has, in general, higher results than the other CRF based classifiers ($p\geq0.20$). 

The results for individual classes in Room3, shown in Figure \ref{Fig:Individual-Rooms} (right column), indicates that, for the minority class Sit-on-bed, dWCRF almost doubles the performance of R-WSVM with statistical significance ($p= 0.018$); and is higher than the other classifiers ($p\geq 0.08$). 
For majority class Lying-on-bed dWCRF is statistically significantly better than fWCRF ($p=0.009$). 
For the minority class Ambulation, dWCRF is lower than R-SVM (4\% difference), but is not statistically significantly different with the other classifiers ($p\geq 0.19$). 
Compared with the other CRF based classifiers dWCRF obtains better results for all classes.

\begin{figure}[tb]
    \centering
    \begin{subfigure}[]{1\columnwidth}
    	\centering
        \includegraphics[width=0.7\linewidth]{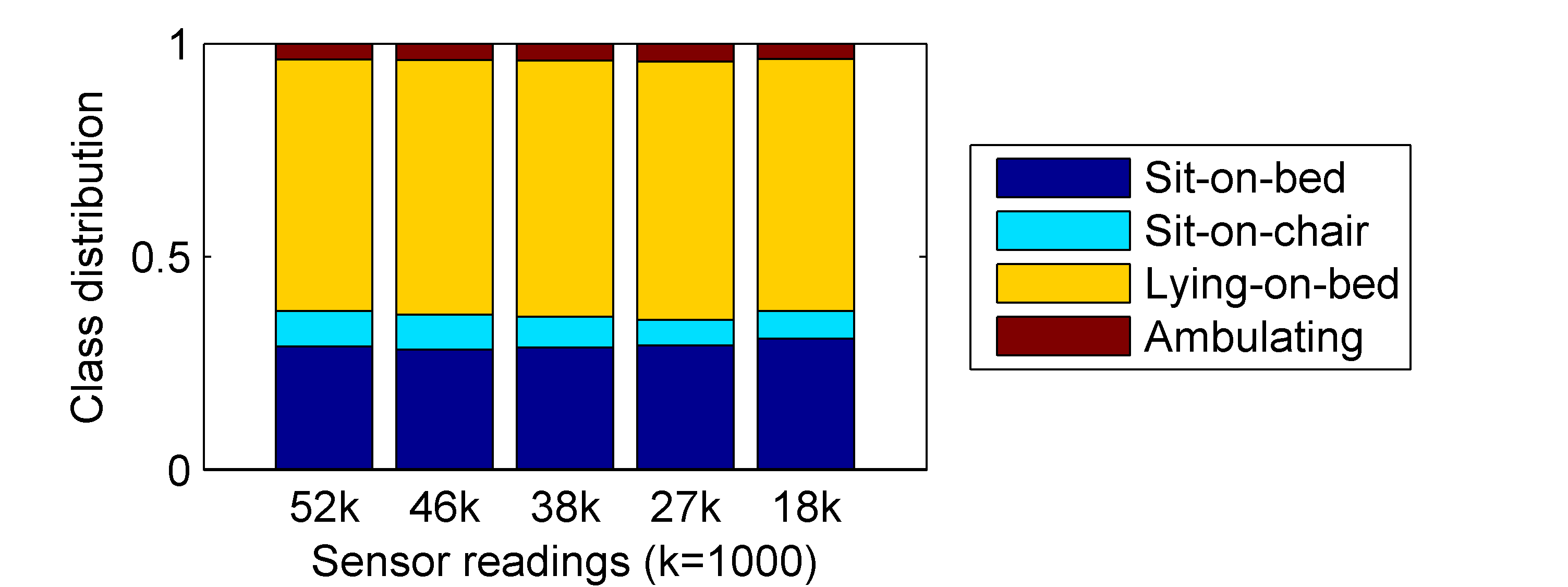}
        \caption{}
        \label{Fig:ReducingDistr}
    \end{subfigure}%
    \\
    \begin{subfigure}[]{1\linewidth}
    	\centering
        \includegraphics[width=0.8\linewidth]{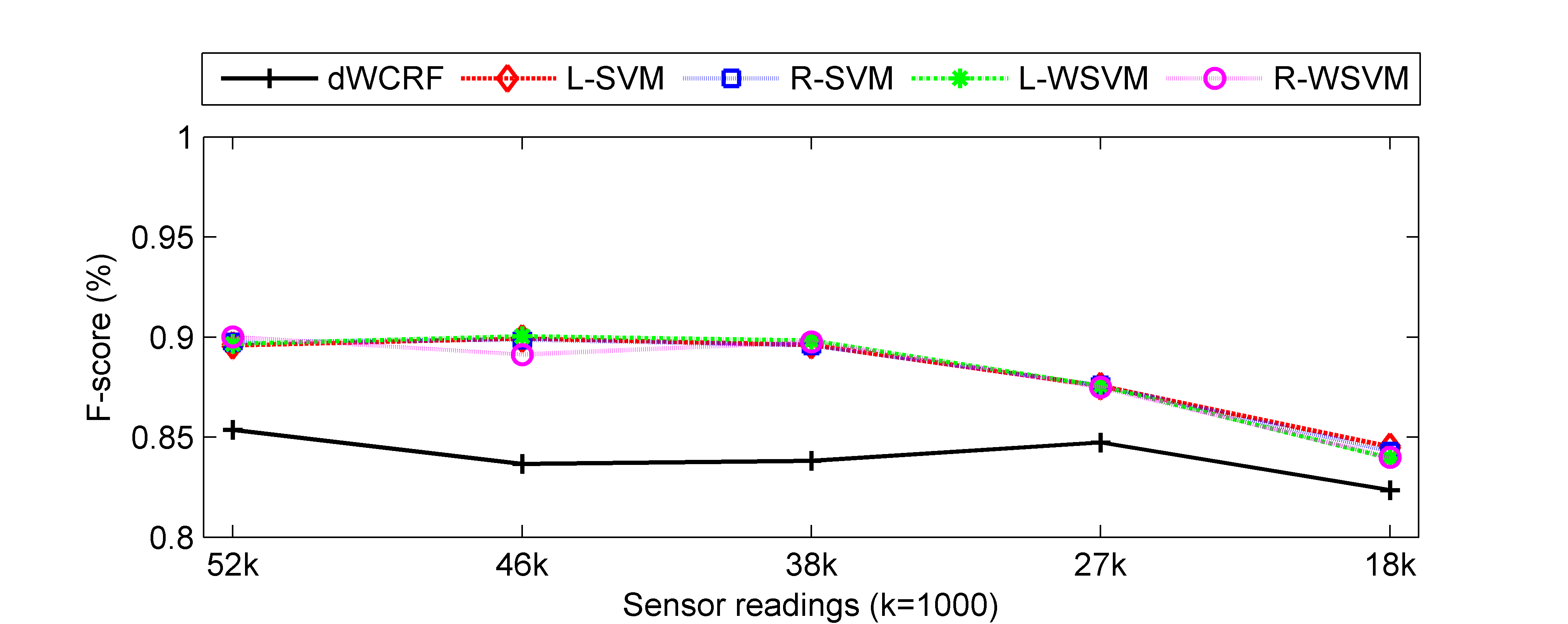}
        \caption{}
        \label{Fig:ReducingRoom1}
    \end{subfigure}%
    \caption{
        (a): Class distribution for different dataset populations showing that classes distributions remain almost unchanged during testing. 
        (b): Results from reducing the population of Room1 dataset to verify its effect on different classifier results. 
    }
    \label{Fig:Reduction}
\end{figure}

In the case of Case Study 3, we consider the performance of the system when learning with limited data. 
For simplicity of dWCRF calculations, we consider parameter $\tau=1$. 
The results from Room1 in Figure~\ref{Fig:Individual-Rooms} (left column) indicate that SVM based algorithms perform statistically significantly better than those of dWCRF ($p\leq0.017$). 
A probable reason is that Room1 contains more than double the information than the other datasets (Section~\ref{ssec:classimb}), providing more than enough support vectors to perform reliable classification. 
We confirm our proposition by experimenting with Room1 dataset by repeatedly reducing one sequence of activities (or trial) from each fold in order to affect each fold evenly.  
This process also does not affect the class distribution of the remaining population 
as illustrated in Figure~\ref{Fig:Reduction}(a). 
Each reduced dataset is tested with the classifiers dWCRF, L-SVM, L-WSVM, R-SVM and R-WSVM. 
The overall performance, shown in Figure \ref{Fig:Reduction}(b) where the \textit{x}-axis displays the approximate number of sensor readings in thousands, indicates that the difference between classifiers reduces as the dataset reduces.

The results in Figure~\ref{Fig:Reduction}(b) indicate that the performance of SVM based classifiers do not vary between each other; moreover, the performance declines after the 38k population marker. 
In contrast, dWCRF performance remains almost unchanged for all population markers. 
The performance difference between SVM classifiers and dWCRF also reduces, we have that between 27k and 18k population markers, this difference is minimal, $\approx4\%$ and $2\%$, respectively. 
More importantly, the differences between classes are no longer statistically significant after the first reduction (46k) with p-values of $p\geq0.51$ for 46k, $p\geq0.56$ for 38k, $p\geq0.41$ for 27k and $p\geq0.44$ for 18k. 

Notice that sensor observation populations, for both Room2 and Room3, are within the population segment between 27k and 18k; and in these datasets, dWCRF results are shown to be similar or superior than all other classifiers. 
This confirms our dWCRF classifier performance being not significantly different than SVM based methods after reducing Room1 population levels to those of Room2 and Room3. 
We also note that Room3 dataset corresponds to a real world scenario with hospitalized patients in a hospital environment. 
Under these real world conditions, we gathered much fewer observations for Room3 than Room1 dataset, although having recruited 25 patients (more than the other two datasets) over a period of more than six months. 
In fact, under most practical situations, collecting and labelling data is difficult and cumbersome. 
Therefore the ability of our dWCRF classifier to learn from imbalanced data when available training data is scarce is a significant result. 

Furthermore, dWCRF has an added advantage over other weighted methods in that dWCRF does not require to search for optimal weights via validation. 
This makes it a faster approach than using a grid search for optimizing the weights or using techniques such as CMA-ES, which requires multiple training initializations while providing no guarantee to achieve optimal results; and where the complexity of optimization increases with the number of classes.

\subsection{Battery Powered Body Worn Sensor Dataset (BPBW)}
\label{ssec:battery}
To further validate the generalizability of our approach, we consider other datasets that showcase sequential information; specifically human activity recognition using wearable sensors. 
Unfortunately, few public datasets on human activity show data imbalance and have enough data samples to modify its levels of imbalance. 
We chose to use the Opportunity activity recognition dataset \cite{Roggen-148362}, available in the UCI repository. 
In this dataset, four participants with multiple sensors attached to their body and the environment perform multiple daily living activities. 
From the multiple sensor data and features generated (243), we only select those that are related to the trunk of the participant, \ie sensors located on the hip and the back of the participant. 
This gives us only 19 sensor related features plus a time related feature. 

These sensors are considered as these are the closest to our real scenario (hospital). 
Moreover, the dataset has modes of locomotion labels: Stand, Walk, Sit, Lie; similarly to the BLBW datasets, the selected torso-based features are used to predict these classes. 
There is also a null class which we consider as an additional class where we assume the participant is doing something other than the four basic locomotion activity labels.  
We also considered observations where readings from both sensors were present, otherwise the observation was discarded. 
Finally, we subdivided the data to 5 datasets, for two main reasons: 
\begin{inparaenum}[i)]
    \item Each dataset contains only one trial from each participant, \ie\ we train and test the system using data from different participants; 
    \item The original set is large $(\approx\num{610000}\text{ observations})$, 
    which is not realistic or convenient 
    since our goal is to evaluate performance with cases of reduced and imbalanced data. 
\end{inparaenum} 
Furthermore, the main objective of this test is not to compare to the established benchmarks, but to compare different methods in situations of high data imbalance not present in the original data.

The original low levels of imbalance are modified for all classes except the original majority class (Stand), in order to create imbalance levels similar and greater than to those of our study cases. 
In this case, we remove readings from the rest of classes on all sequences, \ie data for training and testing have been similarly reduced. 
Three levels of data removal are used as shown in Figure~\ref{Fig:Opp_Results}(a), 
\begin{inparaenum}[Op1:]
    \item remove up to 9 of 10 consecutive sensor readings for each activity;    
    \item remove up to 11 of 12 consecutive sensor readings for each activity; and 
    \item remove up to 14 of 15 consecutive sensor readings for each activity. 
\end{inparaenum}

\subsubsection{Results} 
\label{BPBW:Results}

\begin{figure}[tb]
    \centering
    \begin{subfigure}[]{0.6\linewidth}
        \includegraphics[width=0.9\textwidth]{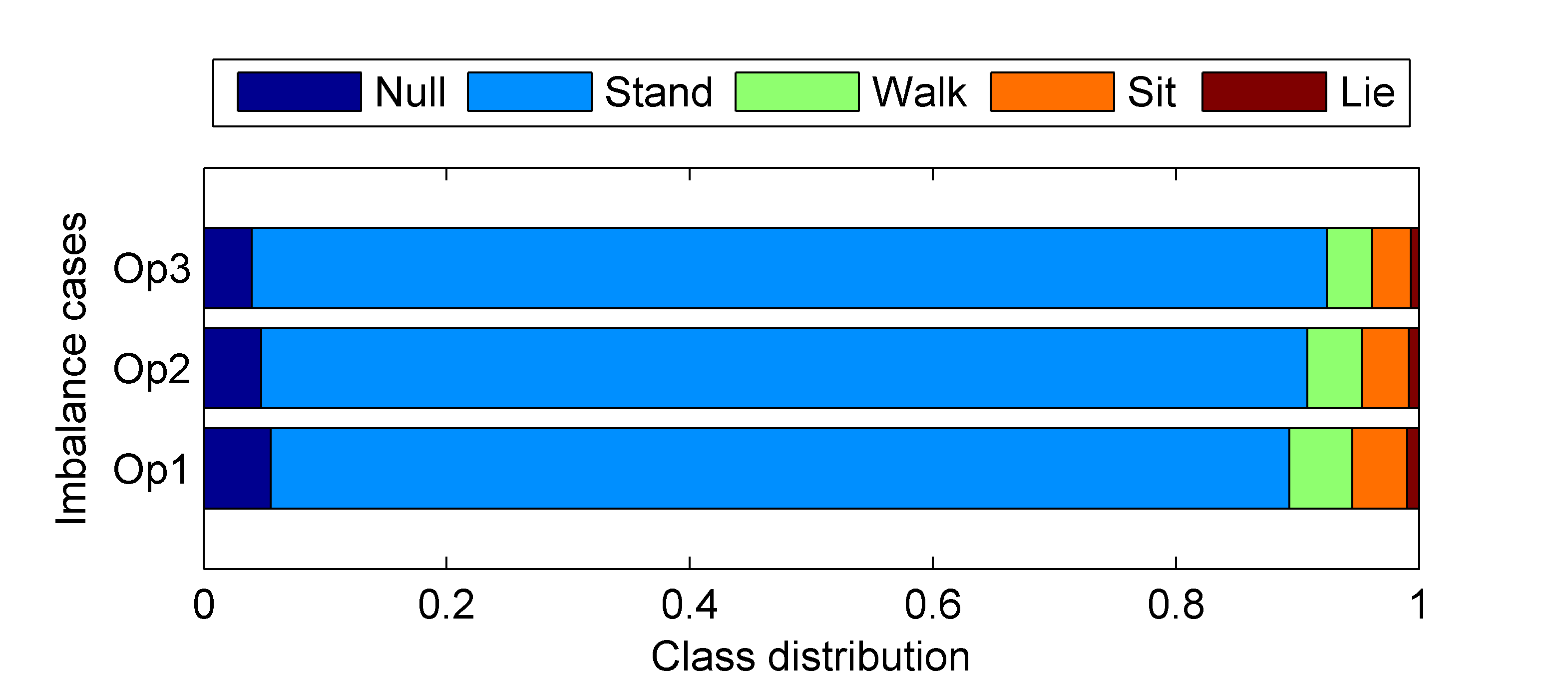}
        \caption{}
        \label{Fig:Opp_Imbalance}
    \end{subfigure}%
    \\
    \begin{subfigure}[]{0.55\linewidth}
        \includegraphics[width=0.9\textwidth]{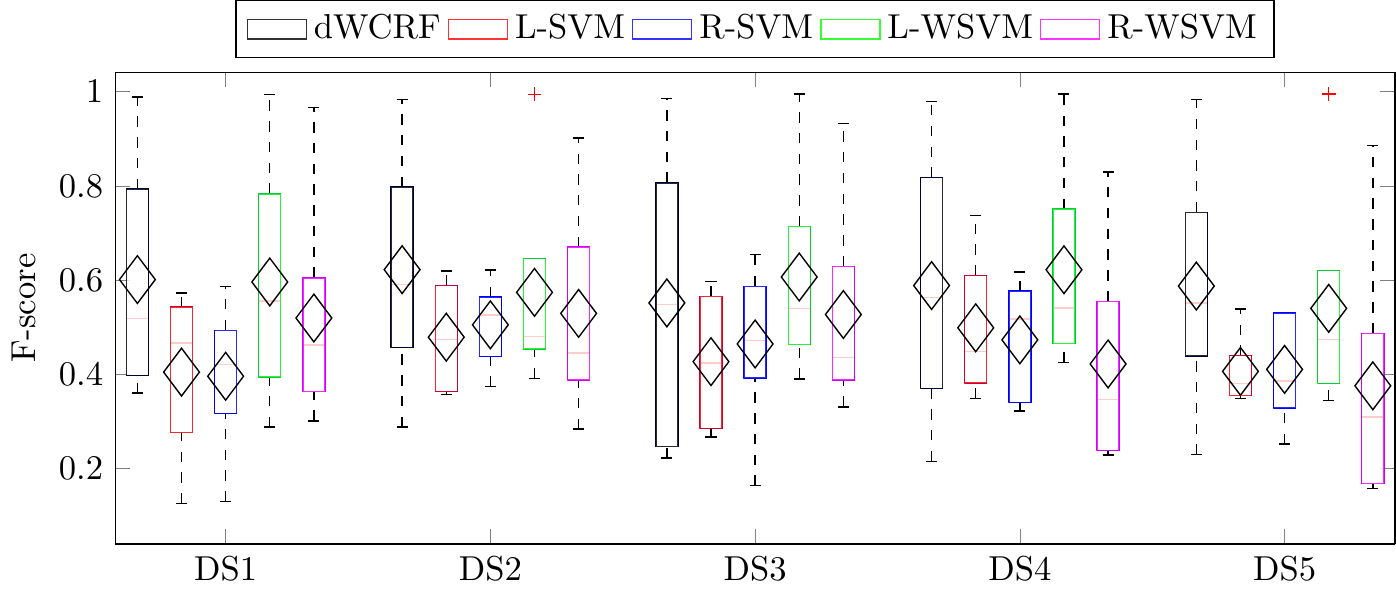}
        \caption{}
        \label{Fig:Opp1}
    \end{subfigure}%
    \\
    \begin{subfigure}[]{0.55\linewidth}
        \includegraphics[width=0.9\textwidth]{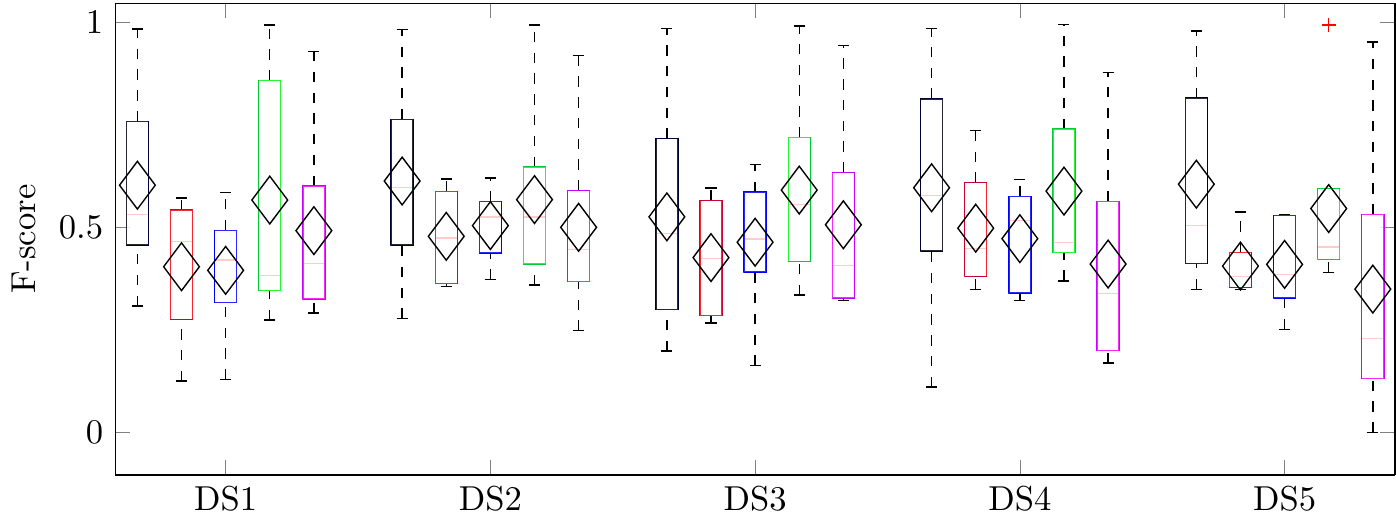}
        \caption{}
        \label{Fig:Opp2}
    \end{subfigure}%
    \\
    \begin{subfigure}[]{0.55\linewidth}
        \includegraphics[width=0.9\textwidth]{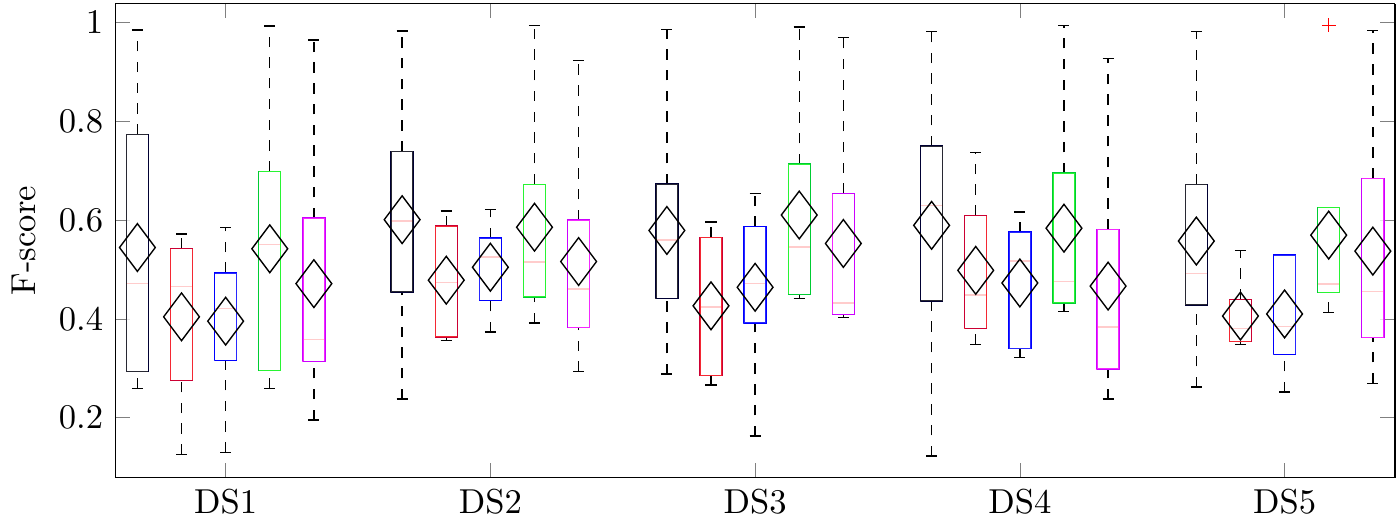}
        \caption{}
        \label{Fig:Opp3}
    \end{subfigure}%
    \caption{(a) Class imbalance distribution over three cases. Overall performance for five datasets, DS1\dots DS5,  on imbalance case (b) Op1, (c) Op2, and (d) Op3.}
    \label{Fig:Opp_Results}
\end{figure}

We test the datasets with the same methods of Case Study 3 in Section~\ref{BLBW:results}. 
the results are shown in Figure~\ref{Fig:Opp_Results}(b)--(d).  
In general, no method is statistically significantly better than the rest with $p\geq0.174$ for Op1, $p\geq0.137$ for Op2 and $p\geq0.251$ for Op3. 
Nonetheless, the average overall \fsc for our dWCRF is in general higher than all other methods, with the exception of L-WSVM, which has higher overall \fsc in 5 (DS3 and DS4 in Op1, DS3 in Op2, and DS3 and DS5 in Op3) of all imbalanced datasets cases. 
Moreover, for dWCRF the lowest class performance in all datasets was for the class Null; whereas for L-WSVM lowest class performance for most datasets was for class Lie, which is the minority class. 
These results suggest that despite high overall results, dWCRF struggles with identifying undefined classes as in these datasets. 
Weights optimization for WSVM methods using CMA-ES required \num{200} initialization processes each dataset (5 datasets) and imbalance case (3 cases); \ie\ \num{3000} CMA-ES processes for each L-WSVM and R-WSVM methods. 
These results validate our method for sequential data with high imbalance and limited training data in terms of performance.

\subsection{Empirical Comparison of Parameter Validation} 

We use the BPBW datasets to illustrate the different training times (including parameter selection) of our proposed dWCRF and other methods. 
From the datasets used in the previous section, we randomly select one dataset from each imbalance case: DS5 from Op1, DS3 from Op2 and DS3 from OP3. 

\begin{figure}[tb]
    \centering
    \includegraphics[width=.5\linewidth]{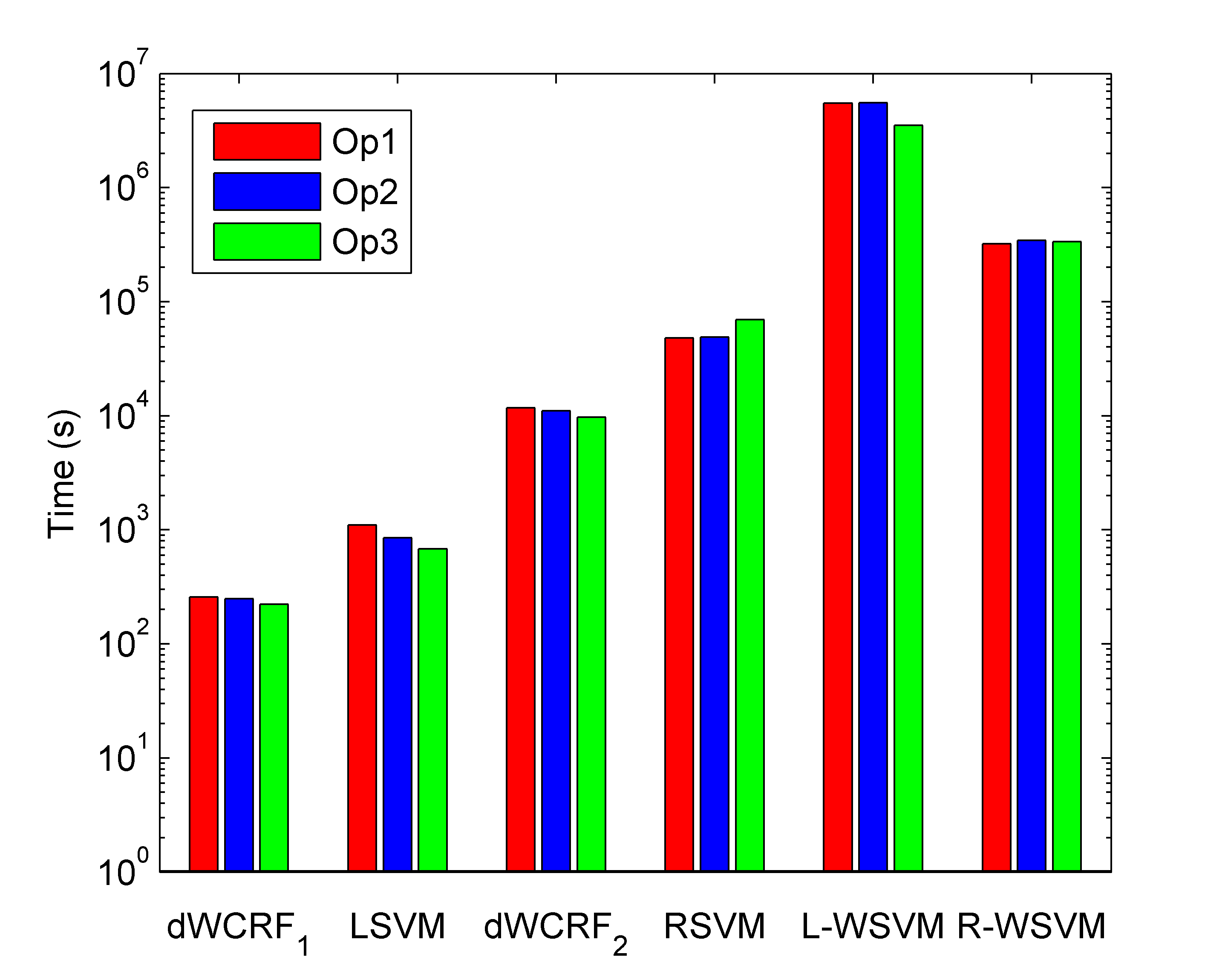}
    \caption{Comparison of validation times for the tested methods: DS5 from Op1 (red), DS3 from Op2 (blue) and DS3 from Op3 (green). Time axis is in logarithmic scale.}
    \label{Fig:Times}
\end{figure}

The resulting times are shown in Figure~\ref{Fig:Times}, where the time axis is in logarithmic scale, where time represents the total time taken to for evaluating the range of validation parameters.  
In this case, we evaluate dWCRF with one parameter to validate (regularization parameter $\theta$ in dWCRF$_1$), and two parameters ($\theta$ and $\tau$ in dWCRF$_2$) with cardinalities of 10 and 121 respectively. 
LSVM had one parameter to validate with a cardinality of 11. 
RSVM had two parameters to validate with a total combination cardinality of 121. 
L-WSVM and R-WSVM added also the times of 200 CMA-ES processes with random starting points, for each method, to validate the values of 5 weight parameters.
It is clear that, increasing or decreasing the number of CMA-ES processes has an impact on the total time; however, a larger number of processes are necessary to obtain a closer to optimal result as a single CMA-ES process does not guarantee an optimal solution. 
We note in Figure~\ref{Fig:Times} the considerable difference between methods and number of parameters; however, the time to validate parameters in dWCRF is lower than that of SVM methods. 
These results confirm the advantages of our method 
delivered in the model selection phase of the training process.

\section{Conclusion}
\label{sec:conclusion}

The present study has established that dWCRF, using a class-wise cost parameter in the objective function, improves the overall \fsc of our batteryless body worn sensor datasets when compared to other CRF based classifiers and performs similar or better than other SVM based classifiers in conditions where training data availability is reduced or collecting large datasets is difficult. 
Moreover, the developed approach also improves \fsc performance metrics on minority classes. 
Our method was validated using a set of battery powered body worn sensor human activity recognition datasets in conditions of high imbalance and limited training data. 

This study also presents a method to obtain a class-wise weighted classifier for optimizing the expected overall \fsc from imbalanced multiclass data. 
In contrast to previous approaches our method dynamically calculates class-wise cost parameters, \ie requires no previous knowledge of the data to assign cost parameter weights and does not need to be evaluated in the validation stage. 
In addition, our method obtained performance results comparable to other cost functions that optimize \fsc such as Softmax-Margin \cite{Gimpel2010} and other SVM based classifiers, but with an extensive reduction in learning time. 
This is because Softmax-Margin function and the weighted SVM methods have more than two parameters to optimize; thus require an exponential amount of validation iterations in order to compare and obtain the best set of parameters. 
The BLBW datasets from old people provide heterogeneous data containing variations in age, health status, physical infrastructure and settings, which were compared using the same features; 
these datasets are scarce as most real-life application data. 
In contrast, most laboratory datasets are well controlled and show little data imbalance which is not representative of real-life conditions. 
In terms of future work, we are interested in testing our approach with other structured classification applications where available data is scarce and imbalanced; as well as investigating the effects of our approach into models of higher order.

\section*{Acknowledgment}

This study was supported by 
the Australian Research Council (DP130104614). 
This study has approval by the Human Research Ethics Committee of the Queen Elizabeth Hospital (protocol number 2011129). 
We also want to thank Prof. Renuka Visvanathan, Mr. Stephen Hoskins and Dr. Shailaja Nair for their support in the selection and supervision of participants for the trials.

\bibliographystyle{ieeetr}
\bibliography{IEEEfull,Bib_ArxivVer}

\end{document}